\documentclass[default,iicol]{sn-jnl}

\usepackage{graphicx}%
\usepackage{multirow}%
\usepackage{amsmath,amssymb,amsfonts}%
\usepackage{amsthm}%
\usepackage{mathrsfs}%
\usepackage[title]{appendix}%
\usepackage{xcolor}%
\usepackage{textcomp}%
\usepackage{manyfoot}%
\usepackage{booktabs}%
\usepackage{algorithm}%
\usepackage{algorithmicx}%
\usepackage{algpseudocode}%
\usepackage{listings}%
\usepackage{tabularray}
\usepackage{xcolor}
\usepackage{changepage}
\usepackage{xurl} 
\usepackage{pifont}  
\newcommand{\ignore}[1]{}

\theoremstyle{thmstyleone}%
\theoremstyle{thmstyletwo}%

\theoremstyle{thmstylethree}%

\begin{document}

\title{AI-driven Remote Facial Skin Hydration and TEWL Assessment from Selfie Images: A Systematic Solution}


\author[1]{\fnm{Cecelia} \sur{Soh}}
\author[1]{\fnm{Rizhao} \sur{Cai}\textsuperscript{\ding{41}}}
\author[2]{\fnm{Monalisha} \sur{Paul}}
\author[1]{\fnm{Dennis} \sur{Sng}}
\author[1]{\fnm{Alex} \sur{Kot}}
 
\affil[1]{Rapid-Rich Object Search (ROSE) Lab, School of Electrical \& Electronic Engineering, Nanyang Technological University (NTU), Singapore}

\affil[2]{Research and Development, Beauty Care, P\&G International Operations(SA), Singapore}
\affil[\ding{41}]{Corresponding author}

\abstract{Skin health and disease resistance are closely linked to the skin barrier function, which protects against environmental factors and water loss. Two key physiological indicators can quantitatively represent this barrier function: skin hydration (SH) and trans-epidermal water loss (TEWL). 
Measurement of SH and TEWL is valuable for the public to monitor skin conditions regularly, diagnose dermatological issues, and personalize their skincare regimens. However, these measurements are not easily accessible to general users unless they visit a dermatology clinic with specialized instruments. 
To tackle this problem, we propose a systematic solution to estimate SH and TEWL from selfie facial images remotely with smartphones. Our solution encompasses multiple stages, including SH/TEWL data collection, data preprocessing, and formulating a novel Skin-Prior Adaptive Vision Transformer model for SH/TEWL regression. Through experiments, we identified the annotation imbalance of the SH/TEWL data and proposed a symmetric-based contrastive regularization to reduce the model bias due to the imbalance effectively. This work is the first study to explore skin assessment from selfie facial images without physical measurements. It bridges the gap between computer vision and skin care research, enabling AI-driven accessible skin analysis for broader real-world applications.}




\maketitle

\section{Introduction}\label{sec1}
Skin barrier function is important in understanding skin health and diseased states. Skin barrier function is closely related to skin hydration (SH)~\cite{correlateshtewl} because proper hydration keeps the skin plump, smooth, and elastic, reducing the appearance of fine lines and wrinkles and giving it a youthful radiance. Besides, the skin's barrier, located in the outermost layer, is crucial in preventing excessive water loss and shielding the skin from environmental stressors~\cite{tewl}. When the barrier is compromised, it can lead to dehydration, increased trans-epidermal water loss (TEWL), irritation, and sensitivity, resulting in redness, rough texture, and dullness. Well-hydrated skin with an intact barrier appears smooth, firm, and radiant, with an even tone, while compromised skin shows visible signs of damage and aging. 
Therefore, hydration and a strong barrier are essential for overall skin health and appearance. 

\textcolor{black}{In general, skin hydration and TEWL are inversely related, which means that as skin hydration increases, TEWL tends to decrease. However, the relationship is not strictly linear~\cite{skinmap}, especially on the facial skin. These two metrics assess two different physiological properties: SH reflects the water content in the stratum corneum, while TEWL reflects the ability to prevent water loss from the skin. A facial skin region may exhibit normal SH levels while still showing a high TEWL. This inconsistency can be attributed to the unique properties of facial skin, which is thinner, more vascular, and more frequently exposed to environmental stressors, leading to fluctuations in TEWL.}

\begin{figure}[t]
\centering
\includegraphics[width=0.48\textwidth]{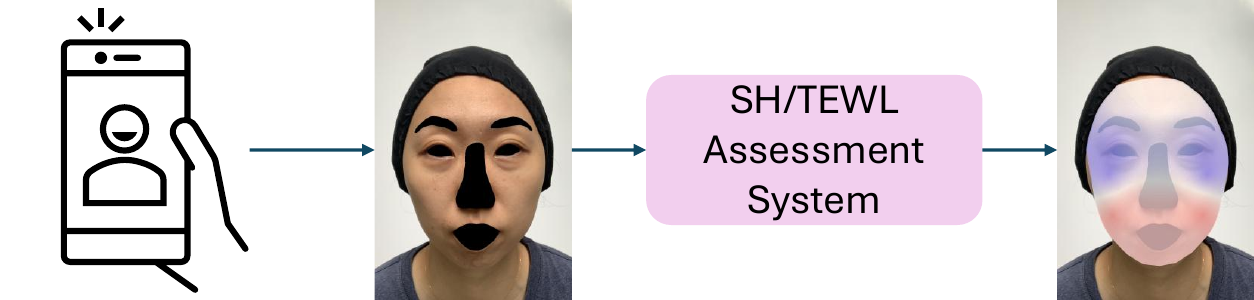}
\caption{Our proposed solution enables the assessment of full-face SH/TEWL on selfie images by generating a visualized heatmap to represent skin conditions.\textsuperscript{1}}
\label{fig:app}
\end{figure}

Measuring SH and TEWL requires expensive, clinical-based tools that are impractical for \textcolor{black}{the average consumer}. For example, SH and TEWL can be assessed using a Corneometer and a VapoMeter, respectively.
However, such devices remain costly and assess only a single point, making full-face analysis time-consuming.
To address these limitations, we proposed an affordable and convenient smartphone-based skin assessment system that enables full-face SH and TEWL evaluations. 
Our system bridges this gap by leveraging visual features, such as skin texture, to estimate SH and TEWL. It then generates heatmaps overlaid on uploaded facial images, with red highlighting areas of concern (poor skin condition) and blue indicating healthier regions, as depicted in Fig. \ref{fig:app}\footnote{\textcolor{black}{Masks are applied to all presented facial images to remove personally identifiable information (PII) and protect participant privacy.}}.

\begin{figure*}[htbp!]
\centering
\includegraphics[width=.92\textwidth]{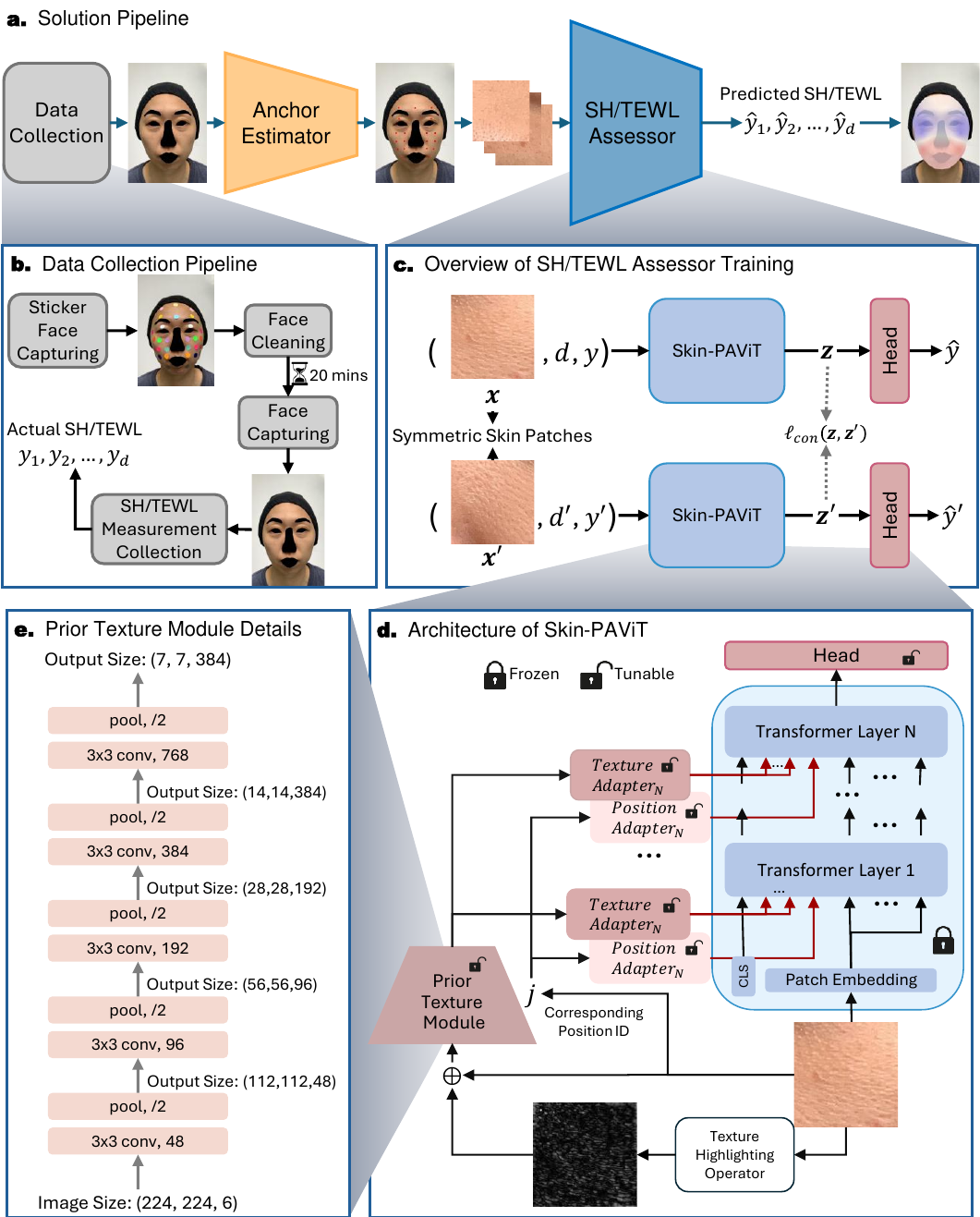}
\caption{\textbf{Solution Overview.}\textsuperscript{1} \textcolor{black}{This figure shows the framework of our solution. \textbf{a}, Our proposed solution includes data collection (Section \ref{sec:data_collect_method}), anchor estimation (Section \ref{sec:centroids}), SH/TEWL prediction (Section \ref{sec:model_arch}) and prediction visualization (Section \ref{sec:heatmap}). \textbf{b}, Each panelist must wash their face before ongoing the facial image and SH/TEWL measurement collection. We also collect the facial image with sticker marking, which aids in estimating SH/TEWL measurements on non-sticker images. The pipeline of measurement anchor estimation is illustrated in Fig. \ref{fig:centroids_model}. \textbf{c}, Overview of the Skin-PAViT training process, which takes pairs of symmetric skin patches and the corresponding position ID $d$ as input, and $y$ as the ground truth value. We apply symmetry-based contrastive learning to increase the similarity of the latent representation. \textbf{d}, Skin-PAViT architecture with a frozen ViT backbone and trainable adapters (Prior Texture Module, Texture Adapters, Position Adapters). \textbf{e}, The detailed architecture of the Prior Texture Module.}
}
\label{fig:pipeline}
\end{figure*}

SH and TEWL are often reflected in the skin's appearance. For example, dry, flaky skin is associated with a low SH~\cite{skinhydration}, while rough, wrinkled skin corresponds to a high TEWL~\cite{poresizeRednessTewl, TDA}. Unlike previous work~\cite{TDA}, which relied on topological data analysis on the skin texture with machine learning technique, to assess the TEWL only on the left cheek \textcolor{black}{images}, we propose an innovative systematic solution that is more generalizable to real-world practical scenarios on a full-face image. By capturing the full-face images, our approach enables the analysis of skin appearance and the assessment of SH/TEWL measurements across various regions. A heatmap is overlaid on facial images to represent the assessment results visually. As shown in Fig.~\ref{fig:pipeline}, our proposed solution includes data collection and processing based on insightful analysis,  skin-specific model architecture design with optimization, and in-depth analysis with visualization. Below, we highlight our contributions in developing this assessment system point-by-point.
\begin{figure}[t]
\centering
\includegraphics[width=0.48\textwidth]{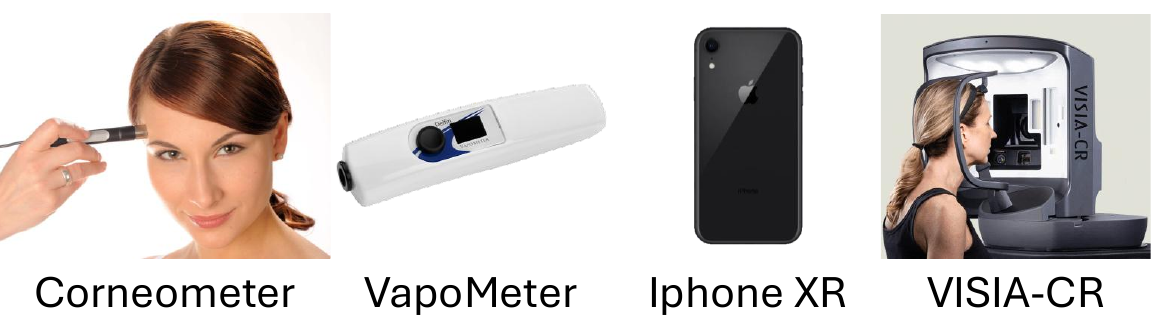}
\caption{Devices used for data collection. We measured the SH and TEWL levels using a Corneometer and VapoMeter, used an iPhone XR to collect selfie images, and collected high-resolution facial images using VISIA-CR.}\label{fig:devices}
\end{figure}

\textbf{Data Collection.} No dataset exists for the remote SH/TEWL estimation. We contribute a practical pipeline of collecting facial images and a scheme of anchor localization to annotate them with SH/TEWL data, measured using a Corneometer\footnote{Produced by Courage \& Khazaka Electronic, Cologne, Germany, \url{https://www.courage-khazaka.com/en/faq?catid=12&id=234&view=article}} and a VapoMeter\footnote{Produced by Delfin, Technologies Ltd, Kuopio, Finland, \url{https://delfintech.com/blog/the-delfin-vapometer-the-benefits-of-closed-chamber-measurement-principle-in-tewl-measurements/}}, as shown in Fig. \ref{fig:devices}.
\textcolor{black}{The ultimate goal of our work is to develop a solution for SH/TEWL assessment using selfie images. However, in real-world scenarios, selfie images may be captured under varying environmental conditions and may contain different types of noise, which could affect the performance of our solution. Therefore, we will also evaluate our approach using facial images captured under controlled conditions, considering them a reference standard for comparison with selfie images.
For that reason, we collected facial images using not only a smartphone (iPhone XR\footnote{Produced by Apple Inc., Cupertino, CA, USA}) but also a VISIA-CR\footnote{Produced by Canfield Scientific, Inc., Parsippany, NJ, USA, \url{https://www.canfieldsci.com/imaging-systems/visia-cr/}}, which offers skin-customized high-resolution images under constrained lighting conditions.
In this paper, the images collected using the two capturing devices are referred to as `Selfie' and `VISIA' images.} 

\begin{figure}[t!]
\centering
\includegraphics[width=0.4\textwidth]{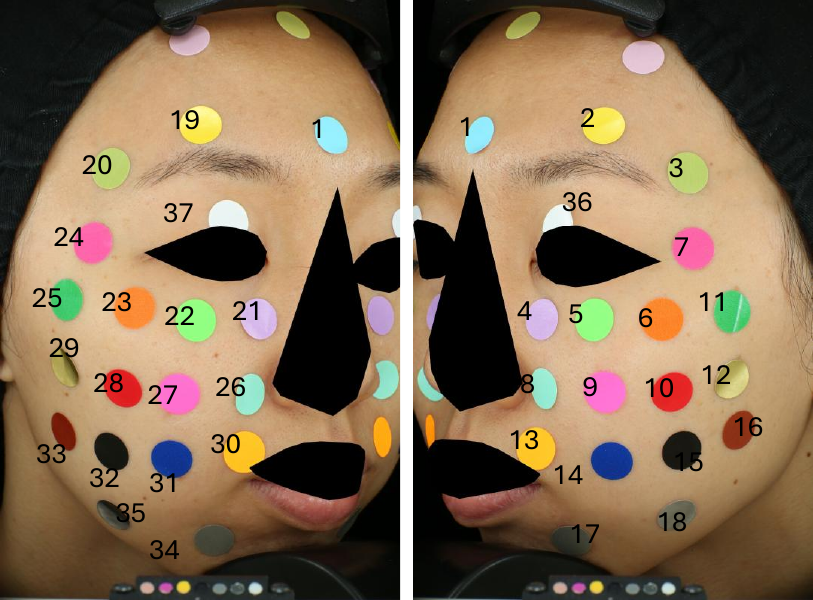}
\caption{The facial position IDs where SH/TEWL measurements were collected.\textsuperscript{1}}\label{fig:data_posid}
\end{figure}

To obtain SH and TEWL annotations, sensors were put on 37 distinct facial points for each subject to ensure comprehensive facial coverage and enable accurate region-specific analysis across the face. Stickers were placed on specific corresponding points to aid in calibration for SH and TEWL estimation from facial images, as shown in Fig.~\ref{fig:data_posid}. Section~\ref{sec:data_collect_method} introduces the data collection procedure, and we discuss the data analysis in Section~\ref{sec:data_collect_analysis}. 
 
\textbf{Data Processing.} To conduct the SH/TEWL assessment from facial images, as shown in Fig. \ref{fig:pipeline}\textbf{a}, we developed an anchor estimator to predict the coordinates of measurement points based on the facial landmarks extracted. \textcolor{black}{Each anchor point defines an image patch, used as a training example with SH/TEWL data annotation.} Section \ref{sec:centroids} presents the detailed pipeline of the anchor estimation, and  Section \ref{sec:result_centroids} discusses the accuracy of this pipeline.

\textbf{SH/TEWL Prediction Model.} We estimate the SH and TEWL from the patch with a neural network model. We proposed an innovative model called Skin-Prior Adaptive Vision Transformer (Skin-PAViT). 
\textcolor{black}{As illustrated in Fig. \ref{fig:pipeline}\textbf{d}, in the training of Skin-PAViT, we initialized the backbone with ImageNet pretrained weights to mitigate underfitting due to limited data. Instead of updating the entire ViT backbone~\cite{vit}, we froze its weights and inserted adapter modules, which are updated during training. These adapters introduce local inductive bias and encourage low-rank feature learning, enabling efficient parameter transfer learning~\cite{evp, vpt, vitAdapter, convpass} for effective model optimization with limited data.}
Compared with vanilla adapter methods, our designed Skin-PAViT consists of several novel trainable modules. \textcolor{black}{Texture Adaptive Module, comprised of a Prior Texture Module and Texture Adapters, takes the high-frequency components from skin patches as input, learning about fine skin texture details such as wrinkles and pores. Additionally, Position Adapters provide the model with positional information about the skin patch on the face.} A more detailed structure of Skin-PAViT is presented in Section \ref{sec:model_arch}, and the ablation study of the effectiveness of different components is shown in Section \ref{sec:result_eff_arc}.

\begin{figure}[t!]
\centering
\includegraphics[width=0.5\textwidth]{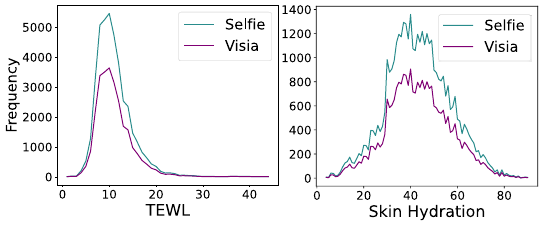}
\caption{The TEWL and SH distribution of the training data for each dataset.}\label{fig:data_dist}
\end{figure}

\begin{table*}[t]
\small
  \centering
  \caption{Comparison of experiments with other models. $\downarrow$ indicates that lower values are better, while $\uparrow$ indicates that higher values are better.}
    \begin{tabular}{|c|l|rrrrr|rrrrr|}
    \hline
       &    & \multicolumn{5}{c|}{TEWL} & \multicolumn{5}{c|}{SH} \\
    \hline
    \multicolumn{1}{|l|}{Dataset} & Methods & \multicolumn{1}{p{2.5em}}{MAE$\downarrow$ (All)} & \multicolumn{1}{p{2.5em}}{MAE$\downarrow$ (Many)} & \multicolumn{1}{p{2.5em}}{MAE$\downarrow$ (Med.)} & \multicolumn{1}{p{2.5em}}{MAE$\downarrow$ (Few)} & \multicolumn{1}{l|}{$R^2$$\uparrow$} & \multicolumn{1}{p{2.5em}}{MAE$\downarrow$ (All)} & \multicolumn{1}{p{2.5em}}{MAE$\downarrow$ (Many)} & \multicolumn{1}{p{2.5em}}{MAE$\downarrow$ (Med.)} & \multicolumn{1}{p{2.5em}}{MAE$\downarrow$ (Few)} & \multicolumn{1}{l|}{$R^2$$\uparrow$} \\
    \hline
    \multirow{9}[2]{*}{Selfie} & ViT-B~\cite{vit} & 2.38 & 1.89 & 2.57 & 8.44 & 0.209 & 9.33 & 6.31 & 16.66 & 26.78 & 0.133 \\
       & ResNet-18~\cite{resnet} & 2.58 & 2.10 & 2.63 & 8.65 & 0.111 & 10.29 & 7.37 & 17.30 & 26.73 & -0.044 \\
       & ResNet-50~\cite{resnet} & 2.47 & 1.96 & 2.65 & 8.35 & 0.177 & 9.92 & 7.07 & 16.76 & 26.71 & 0.018 \\
       & VGG-16~\cite{vgg} & 2.42 & 1.96 & 2.53 & 8.63 & 0.196 & 9.63 & 6.23 & 18.10 & 28.44 & 0.070 \\
       & EffNet-B0~\cite{effnet} & 2.49 & 1.97 & 2.64 & 8.31 & 0.156 & 9.85 & 6.92 & 16.96 & 26.98 & 0.037 \\
       & ConvNeXt-B~\cite{convnext} & 2.35 & 1.75 & 2.65 & 8.58 & 0.210 & 9.50 & 6.28 & 17.35 & 26.55 & 0.099 \\
       & Swin-B~\cite{swin} & 2.30 & 1.71 & 2.62 & 8.53 & 0.236 & 9.40 & \textbf{6.19} & 17.01 & 26.67 & 0.118 \\
       & PVT-M~\cite{pvt} & 2.38 & 1.95 & 2.45 & 8.65 & 0.212 & 9.43 & 6.53 & 16.16 & 25.19 & 0.110 \\
       & \textbf{Ours} & \textbf{2.28} & \textbf{1.62} & \textbf{2.41} & \textbf{6.49} & \textbf{0.264} & \textbf{8.97} & 6.33 & \textbf{15.04} & \textbf{24.92} & \textbf{0.205} \\
    \hline
    \multirow{9}[2]{*}{VISIA} & ViT-B~\cite{vit} & \textbf{2.19} & 1.80 & 2.42 & 7.99 & 0.318 & 9.21 & 6.54 & 14.92 & 23.81 & 0.187 \\
       & ResNet-18~\cite{resnet} & 2.35 & 1.83 & 2.78 & 7.76 & 0.242 & 9.66 & 6.38 & 17.11 & 27.61 & 0.112 \\
       & ResNet-50~\cite{resnet} & 2.26 & 1.79 & 2.49 & 7.88 & 0.303 & 9.57 & 6.38 & 16.77 & 27.11 & 0.124 \\
       & VGG-16~\cite{vgg} & 2.22 & 1.74 & 2.55 & 8.73 & 0.307 & 9.65 & 6.70 & \textbf{13.93} & 24.17 & 0.118 \\
       & EffNet-B0~\cite{effnet} & 2.38 & 1.79 & 2.72 & 8.71 & 0.203 & 9.80 & 7.11 & 15.64 & 24.12 & 0.094 \\
       & ConvNeXt-B~\cite{convnext} & 2.29 & 2.04 & 2.19 & 7.79 & 0.301 & 9.49 & 6.81 & 15.35 & 25.77 & 0.149 \\
       & Swin-B~\cite{swin} & 2.22 & 1.81 & 2.45 & 7.67 & 0.307 & 9.21 & 6.54 & 14.92 & 23.81 & 0.187 \\
       & PVT-M~\cite{pvt} & 2.20 & \textbf{1.79} & 2.30 & 8.10 & 0.315 & 9.12 & \textbf{6.27} & 15.34 & 23.55 & 0.212 \\
       & \textbf{Ours} & \textbf{2.19} & 1.86 & \textbf{1.99} & \textbf{7.67} & \textbf{0.348} & \textbf{8.83} & 6.45 & 14.10 & \textbf{22.25} & \textbf{0.263} \\
    \hline
    \end{tabular}%
  \label{tab:results}%
\end{table*}%

\textbf{Cross-lighting Generalization.} Skin tone is correlated with skin hydration~\cite{skinmap}, while facial redness increases  TEWL~\cite{poresizeRednessTewl}. \textcolor{black}{However, the skin tone and redness captured in images can be distorted in hue and saturation because of the influence of environmental lighting. Considering how lighting affects skin coloration in images is important for understanding its impact on SH/TEWL predictions.} During the data collection process, we collected the facial images under various lighting conditions. Nevertheless, our dataset's diversity of lighting scenarios remains limited, and collecting more data is labor-intensive and time-consuming. To cope with this issue, we introduced a lighting data augmentation technique that modified the skin coloration presentation in images to enhance the model's robustness against different lighting environments. We evaluated the effectiveness of our method through a leave-one-lighting-out experiment on the Selfie dataset, and the results are presented and discussed in Section \ref{sec:result_lighting}. The detailed lighting augmentation method is described in Section \ref{sec:lighting_aug}.

\begin{figure*}[t!]
\centering
\includegraphics[width=0.85\textwidth]{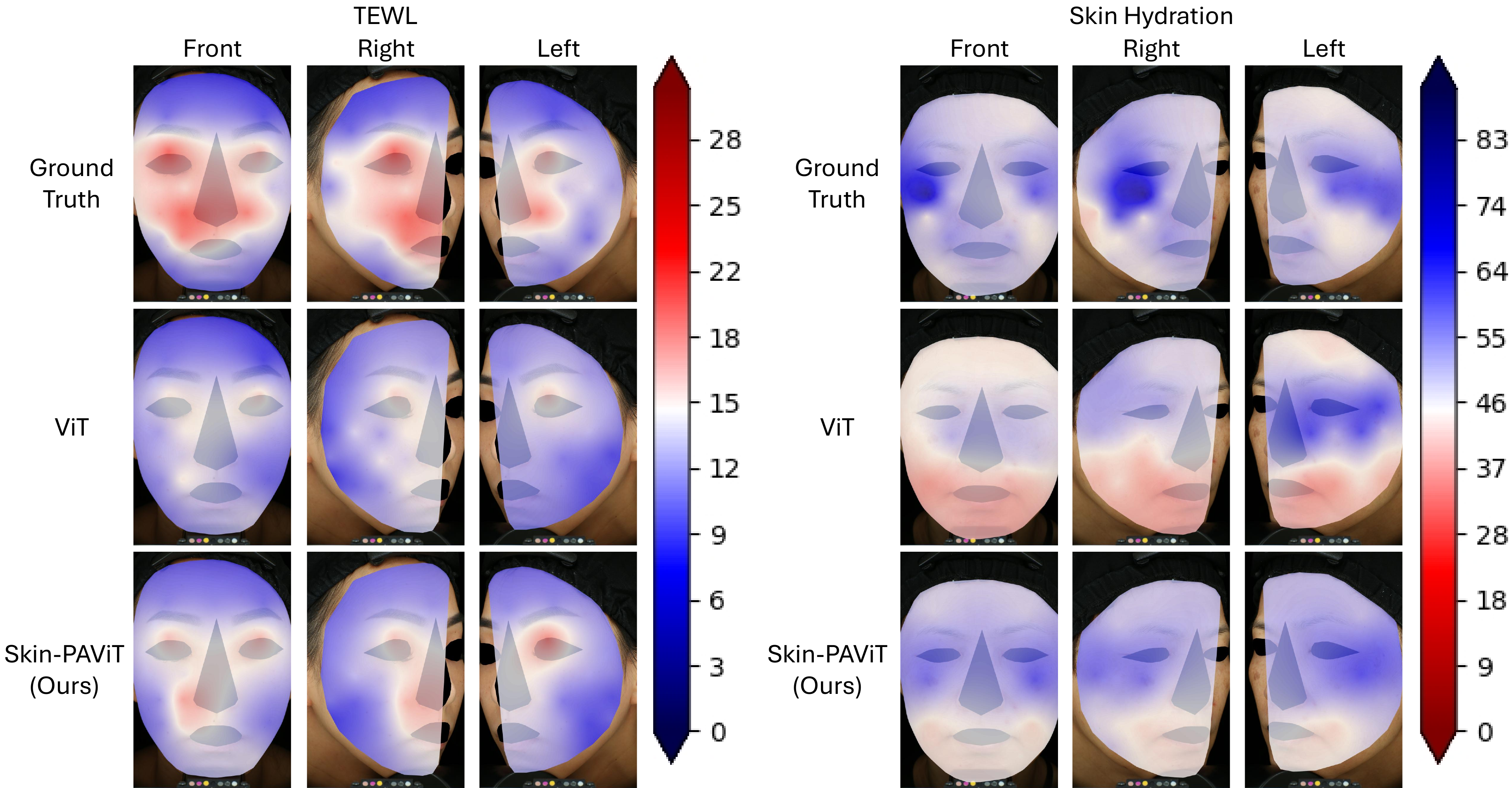}
\caption{\textcolor{black}{\textbf{Heatmap Visualization on VISIA Images.}\textsuperscript{1} The presented subjects have more SH/TEWL measurements that fall within a minor range, meaning their ground truth heatmaps differ significantly from the average heatmap (Fig.~\ref{fig:data_ave_heatmap}.). Our method is able to generate heatmaps that more closely resemble the ground truth, whereas the fully fine-tuned ViT tends to produce heatmaps that resemble the average heatmap, as shown in Fig.~\ref{fig:data_ave_heatmap}.}}\label{fig:results_heatmap}
\end{figure*}

\begin{figure*}[t!]
\centering
\includegraphics[width=0.85\textwidth]{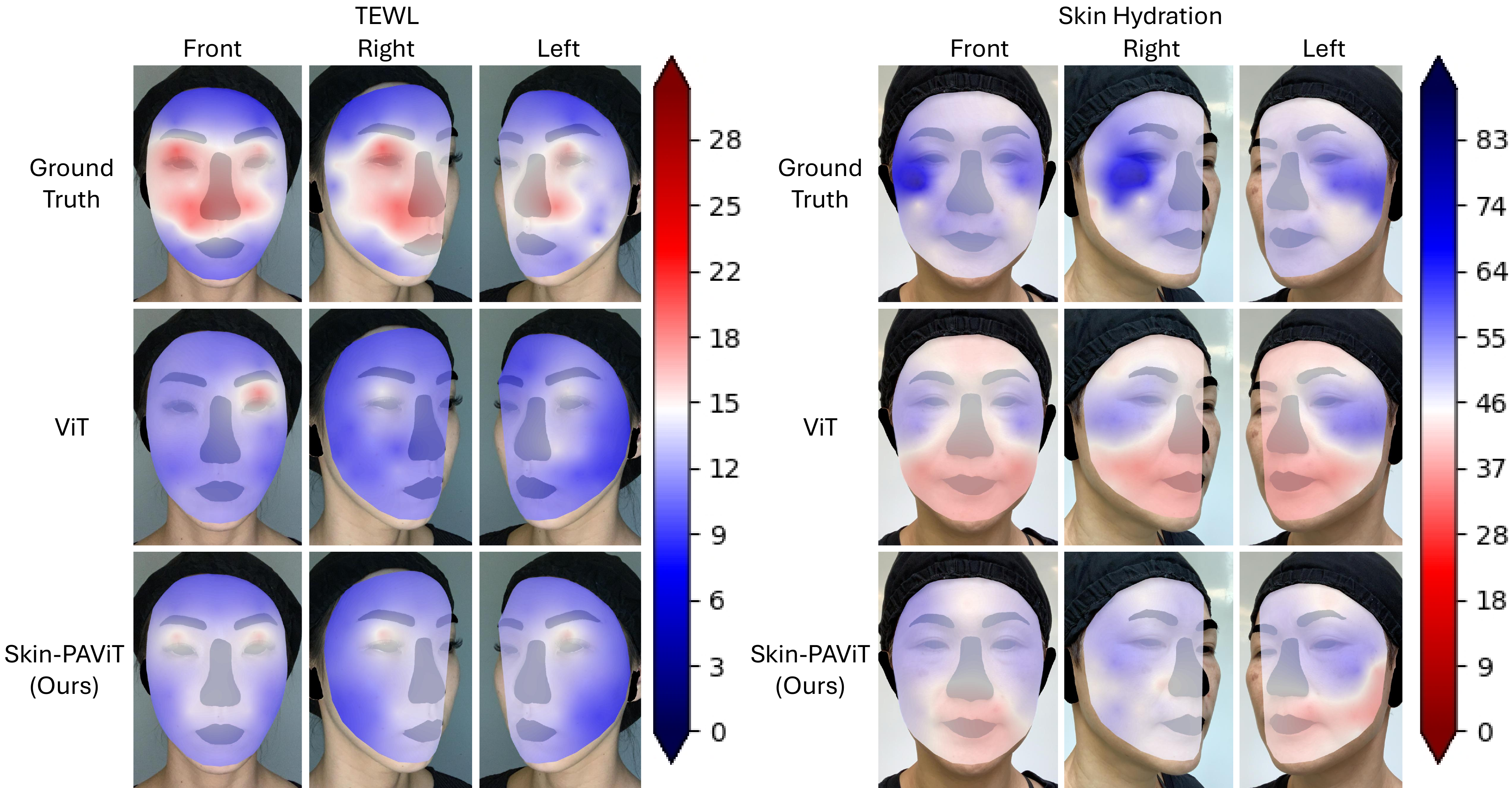}
\caption{\textcolor{black}{\textbf{Heatmap Visualization on Selfie Images.}\textsuperscript{1} The presented subjects are the same with Fig.~\ref{fig:results_heatmap}. Similarly, our method is able to generate heatmaps that are more similar to the ground truth than those generated by the fully fine-tuned ViT.}}\label{fig:results_heatmap_selfie}
\end{figure*}

\textbf{Imbalance Regression.} Our application aims to assess as many facial regions as possible, but certain areas, such as the face's contour, tend to be well-hydrated, resulting in an imbalanced distribution of SH/TEWL values, as illustrated in Fig.~\ref {fig:data_dist}. 
This imbalance can bias the model toward the dominant range of SH/TEWL values, making it struggle to recognize samples from the minor range and predict them based on patterns learned from the dominant range. 
To tackle this problem, we employed data augmentation (Section~\ref{sec:DA}) during training and introduced a facial symmetry-based contrastive learning approach (Section~\ref{sec:sym_con}). We also prove the effectiveness of these methods in Section~\ref{sec:result_eff_arc}.


\textbf{Heatmap Visualization.} 
\textcolor{black}{To represent skin condition effectively, we designed the heatmap range based on the typical SH and TEWL values measured by Corneometer and VapoMeter in clinical and cosmetic studies~\cite{corneo_metrics, vapo_metrics}. For SH, normal skin (40–50AU) appears white, dry skin ($<$ 40AU) red, and well-hydrated skin ($>$ 50AU) blue, where AU denotes arbitrary units measured by the Corneometer. TEWL measurements are expressed in grams per square meter per hour ( $g\cdot m^{-2}\cdot h^{-1}$), which reflects the amount of water lost through the skin over a specific area and time. For TEWL measurement, healthy skin ($< 15 g\cdot m^{-2}\cdot h^{-1}$) is shown in blue, while higher values indicating barrier damage are shown in red. The detailed heatmap generation process is described in Section~\ref{sec:heatmap}.
}

Eventually, as shown in Table \ref{tab:results}, our method outperforms other CNN and ViT models, achieving prediction performance  $R^2=0.264$ for TEWL and $R^2=0.205$ for SH on Selfie images and $R^2=0.348$ and $R^2=0.263$, respectively, on VISIA images. The assessment results from these skin patches are then aggregated to generate a heatmap corresponding to the predicted anchors. As displayed in Fig. \ref{fig:results_heatmap} \textcolor{black}{and  Fig. \ref{fig:results_heatmap_selfie}}, our models can generate a similar heatmap for the subjects with more skin measurements within the minor range.
\textcolor{black}{In summary, our contribution includes collecting a comprehensive dataset and designing a novel skin-specific model architecture and training technique for estimating SH and TEWL from facial images, allowing for broader skin condition assessment across various facial regions.}


\section{Background and Related Work}\label{sec:background}
\subsection{SH/TEWL Measurement}\label{sec:measurement}
\textcolor{black}{
Skin hydration (SH)~\cite{skinhydration, corneo_metrics} refers to the water content present in the stratum corneum, i.e., the outermost layer of the skin. Adequate SH is crucial for maintaining barrier integrity, elasticity, and overall skin health. SH is quantitatively assessed using a Corneometer\textsuperscript{2}, a non-invasive device that measures the moisture content of the stratum corneum. The Corneometer operates based on the principle of capacitance, where the device applies a small electrical field to the skin surface. The capacitance changes are directly correlated with the water content in the stratum corneum, allowing for an accurate assessment of skin hydration levels. The Corneometer measures SH in arbitrary units (AU)~\cite{corneo_metrics}.
}

\textcolor{black}{
Trans-epidermal water loss (TEWL)~\cite{tewl, vapo_metrics} quantifies the amount of water vapor that diffuses through the skin into the environment, providing a direct indication of the skin’s barrier efficacy. This assessment is typically conducted using a VapoMeter\textsuperscript{3}. The VapoMeter operates by measuring the water vapor concentration gradient above the skin surface and calculating the rate of water loss. Elevated TEWL rates are indicative of a compromised barrier, often associated with skin conditions such as eczema, psoriasis, or dermatitis. The VapoMeter allows researchers to evaluate the effectiveness of topical treatments, moisturizers, or cosmetic formulations by monitoring changes in TEWL before and after application. TEWL measurements are typically expressed in grams per square meter per hour ($g\cdot m^{-2}\cdot h^{-1}$)~\cite{vapo_metrics}, offering a standardized metric for comparing barrier function between different types of skin and conditions.
}

In addition to Corneometer and VapoMeter, a portable GPSKIN Barrier\footnote{Produced by GPOWER Inc., South Korea, \url{https://mygpskin.com/}} integrates both SH and TEWL measurements into a single device, reducing the need for multiple instruments for the assessment of two metrics. However, all these devices are invaluable tools in both clinical and cosmetic research, but they are not commonly used in domestic settings. They can only measure a single point at a time, making it difficult to assess the SH/TEWL of the entire face simultaneously. This results in an inconvenient and time-consuming process.

While existing applications like ModiFace\footnote{Developed by ModiFace Inc., Toronto, Canada, \url{https://modiface.com/}} can analyze skin surface appearance through selfies, no open solutions currently exist for assessing SH or TEWL from such images.
The only open machine learning solution~\cite{TDA} predicts TEWL by taking the cheeks' skin images as input. It studied the correlation between the TEWL and skin texture using topological data analysis. However, this method is limited in scope as it does not generalize well to the entire face. Our proposed method is more generalizable and comprehensive, and utilizes deep learning techniques to predict the SH/TEWL from a full-face image. This makes the assessment more accessible and practical for real-world applications.

\subsection{Vision Transformer}\label{sec:vit}

Transformer~\cite{transformer} architecture has been widely used to develop Language Models (LMs)~\cite{bert, t5, gpt3, llama} for handling Natural Language Processing (NLP) tasks. Recently, Vision Transformer (ViT)~\cite{vit} has also been widely used for Computer Vision tasks. Derived from the tokenization of word sequences in Transformer models, ViT tokenizes the input images by dividing the images into multiple fixed-size patches. Thanks to the Multi-Head Self-Attention (MHSA) mechanism, ViT is able to learn about global relations by capturing long-range dependencies among patches. This capability is not possessed by conventional CNN, helping ViT achieve state-of-the-art performance in many computer vision tasks~\cite{vit, pvt, swin, mvitv2}, such as image classification, object detection, and segmentation.  Leveraging a pretrained ViT backbone allows us to utilize the powerful visual feature extraction capability. Thus, we adopt ViT pretrained from ImageNet~\cite{imagenet} as the backbone of our approach and adapt it from a classification setting for regression. However, ViT is less capable of learning local feature dependencies than CNNs, while capturing subtle local features from skin images is crucial for hydration estimation. To overcome this limitation, our proposed method integrates local feature learning with ViT to compensate for its weakness while still retaining the strength of ViT.

\textcolor{black}{
Moreover, since our dataset is relatively small, we also pay attention to fine-tuning ViT effectively under limited data. We explore Parameter-Efficient Fine-Tuning (PEFT) techniques, which originated from tackling NLP downstream tasks.} These techniques adapt the language models, which have been pretrained on extensive corpora, for the downstream tasks by adding trainable adapter modules to each layer~\cite{nlp-adapter, nlp-parallel_adapter, nlp-lora} or the embeddings to input sequences~\cite{nlp-prefix, nlp-prompt_prepend, nlp-p-tunning, nlp-p-tunning2, nlp-llama-adapter}. In computer vision, several studies~\cite{vpt, vitAdapter, convpass, evp} proposed PEFT methods for ViT models in order to efficiently adapt the model to downstream tasks. For example, VPT~\cite{vpt} leverages prompt-based learning from NLP. During training, only the prompts added to the input sequence are learned for the downstream task. ViT-Adapter~\cite{vitAdapter} enhances inductive biases by introducing adapter modules for dense prediction tasks. EVP~\cite{evp} proposed learning the prompts based on explicit visual content to solve low-level structure segmentation tasks. For other subtle visual learning tasks (face spoofing or deepfake detection), the PEFT also show significant benefits \cite{C1,C2,C3,C4,C5,C6}.

\textcolor{black}{Those existing adaptive models are designed for classical computer vision tasks, such as object classification, detection and segmentation. Distinct from these methods, our proposed adaptive ViT is tailored explicitly for regression problems. Nonetheless, our approach is inspired by these earlier methods.} We freeze our ViT backbone and only train our proposed adapter modules, reducing the number of training parameters \textcolor{black}{while maintaining superior performance in extracting relevant skin feature information, which is beneficial for the prediction.}

\begin{figure}[t!]
\centering
\includegraphics[width=0.40\textwidth]{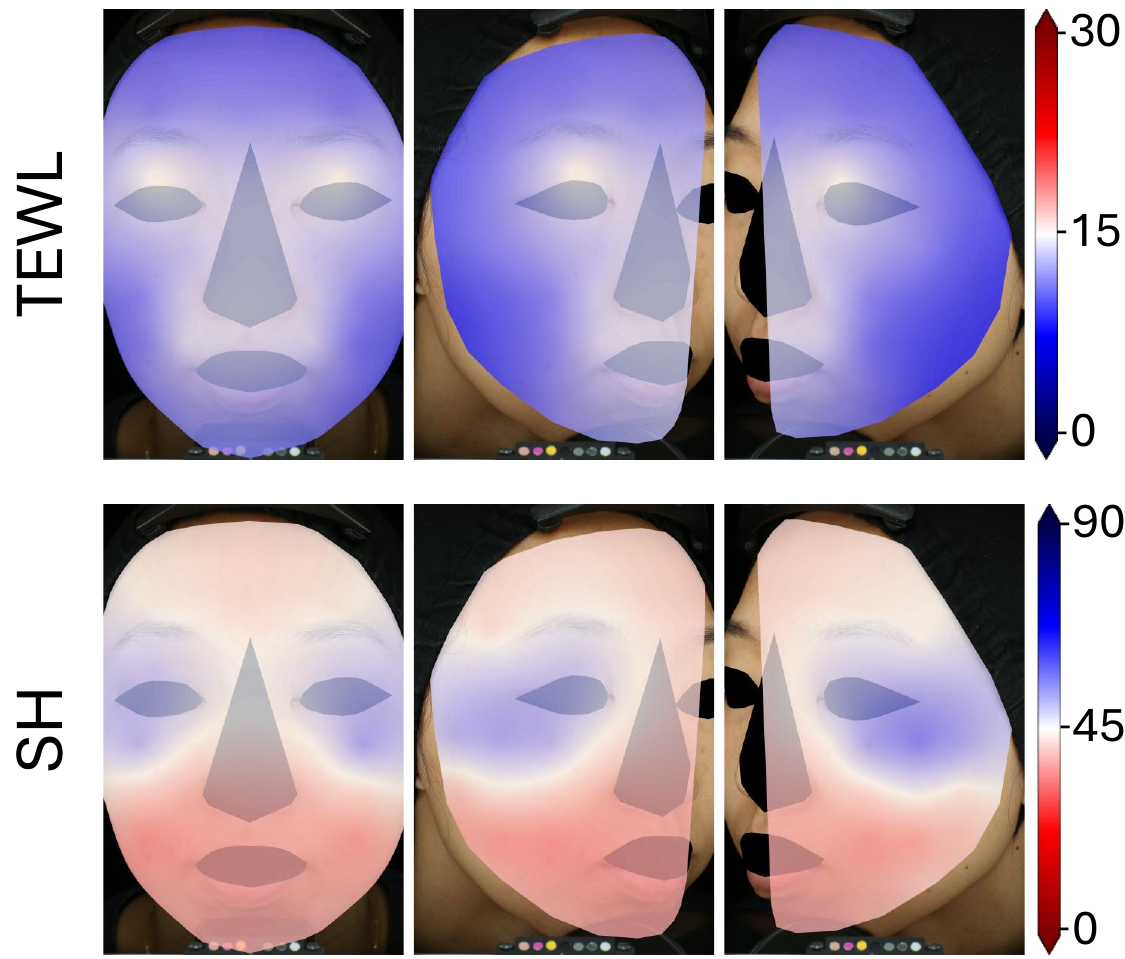}
\caption{These heatmaps were generated from each position's average SH/TEWL values.\textsuperscript{1}}\label{fig:data_ave_heatmap}
\end{figure}

\subsection{Imbalanced Regression}\label{sec:imbalanced}
\textcolor{black}{
Compared to the imbalanced classification problem, the issue of imbalance in regression tasks remains relatively underexplored. Earlier approaches primarily focused on data-level strategies to address this challenge. For example, \cite{torgo2013smote, branco2017smogn} proposed oversampling techniques that generate synthetic data in label regions with fewer samples, while \cite{branco2018rebagg} introduced a resampling-based ensemble method to improve prediction performance in underrepresented label ranges. Over the past few years, \cite{steininger2021density, yang2021delving, ren2022balanced} proposed rebalancing the loss function based on the label distribution. There are also some studies~\cite{yang2021delving, gong2022ranksim} that solve the imbalanced regression problem from a feature perspective by considering the distances among the labels of the training data. The most recent study~\cite{wang2023variational} introduces a novel probabilistic deep learning framework to tackle the challenges by enhancing both prediction accuracy and uncertainty estimation.
}

\textcolor{black}{
Existing methods address the imbalanced regression problem by placing greater emphasis on samples from underrepresented (minority) label ranges, thereby indirectly reducing the influence of samples from overrepresented (majority) ranges.
However, our task differs from traditional regression, which typically aims to predict a target value given an input $x$, as formulated by $\hat{y} = f(x)$. 
In our case, the label distribution can be viewed as an aggregation of multiple sub-distributions corresponding to different facial regions, represented as $\hat{y} = f(x, d)$, where $d$ denotes the region.
Fig.~\ref{fig:data_ave_heatmap} presents heatmaps of the average SH/TEWL measurement for each region. The majority range of some specific regions may fall into the minority range of the overall distribution. For example, in the average TEWL heatmap, the eyelids typically exhibit higher values than the overall majority range. If existing methods are applied directly, the model may overemphasize the eyelid region while overlooking areas like the cheeks, which users care about more. For this reason, we propose a moderate approach to address the imbalanced regression problem by leveraging the symmetrical properties of the face.
}

\begin{figure}[t!]
\centering
\includegraphics[width=0.5\textwidth]{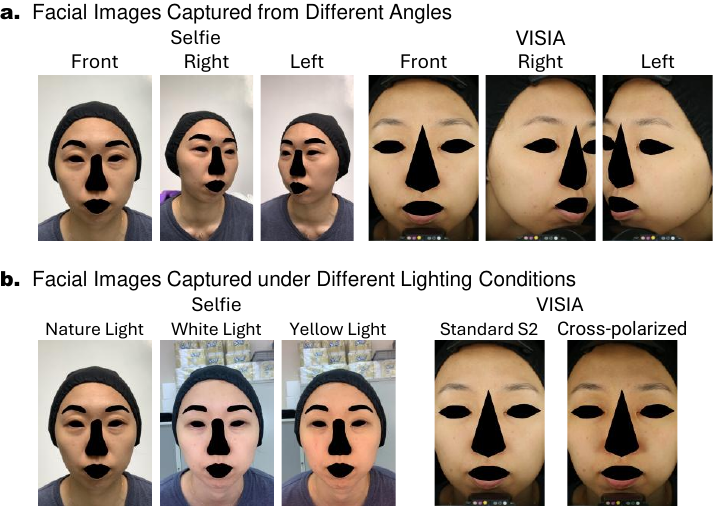}
\caption{\textbf{Dataset Variety.}\textsuperscript{1} \textbf{a}, For each panelist, we collect facial images from three angles using a smartphone and VISIA-CR. \textbf{b}, We also collect the images under different lighting conditions.}\label{fig:data}
\end{figure}

\section{Method}\label{sec:method}
To design an SH/TEWL estimation system using selfie images, we need a high-quality and comprehensive dataset to train the model and ensure its accuracy. Unfortunately, no suitable datasets have been published. Therefore, in this section, we begin by presenting our dataset collection process, followed by the method to identify and crop the skin patches for SH/TEWL assessment. Next, we introduce our proposed Skin-PAViT model for estimating SH and TEWL. Additionally, we present our solutions for handling lighting generalization problems and addressing the imbalance regression issues. At the end of this section, we demonstrate the heatmap generation method, showcasing the prediction results for human faces.

\begin{figure*}[t!]
\centering
\includegraphics[width=0.75\textwidth]{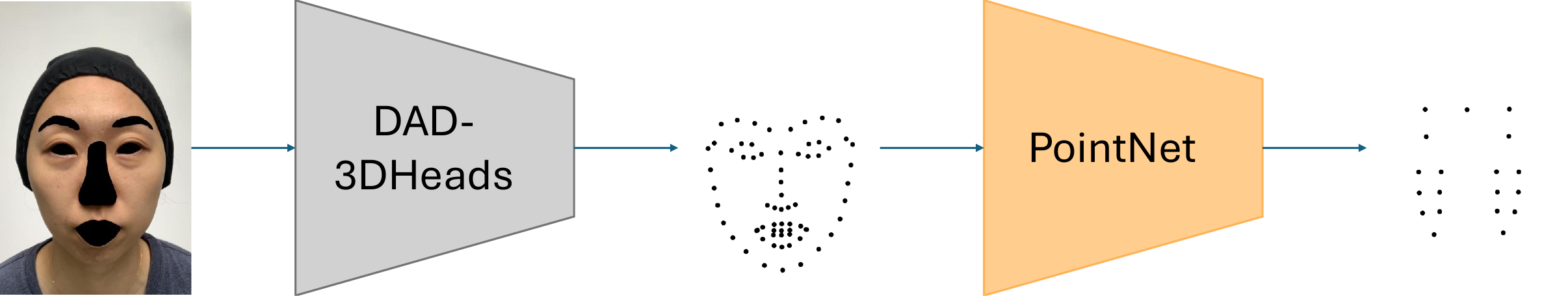}
\caption{We train a PointNet, which takes the facial landmarks generated by DAD-3DHeads as input, to estimate the coordinates of the anchors.\textsuperscript{1}}\label{fig:centroids_model}
\end{figure*}

\subsection{Data Collection Strategy}\label{sec:data_collect_method}

As presented in Fig. \ref{fig:pipeline}\textbf{b}, we first collected the facial images with the stickers placed on key facial regions to mark measurement points. 
After this, panelists were instructed to wash their faces to remove any surface contaminants. After waiting 20 minutes, we captured the panelists' facial images and collected the SH and TEWL measurements using the Corneometer (Model CM825) and  VapoMeter (Model SWL5001JT), respectively, in succession.

We collected two kinds of facial images: regular Selfie images using iPhone XR and high-resolution \textcolor{black}{VISIA} facial images using the VISIA-CR system, which can capture the facial images under various controlled lighting modalities. We aim to study our method not only for general selfie images, which may be captured in various environments, particularly with respect to lighting, but also for facial images taken in a controlled environment.


At the same time, we captured panelists' faces from multiple angles, including side views, which reveal the facial contours and provide a complete representation. As displayed in Fig. \ref{fig:data}\textbf{a}, we collected the Selfie and VISIA images from multiple facial angles. This is due to our model being designed to analyse skin patches across the entire face. 

We accounted for skin coloration under various lighting conditions by capturing facial images with different light sources. As presented in Fig. \ref{fig:data}\textbf{b}, Selfie images were taken under natural, white, and yellow light, resulting in distinct skin coloration representations. In addition, wrinkles captured under natural light are more obvious than those captured with a light source illuminating the face. This comes out in the variety of image captures that occur in real-world scenarios. 
On the other hand, we collected images under two controlled lighting modalities using the \textcolor{black}{VISIA-CR} system: `Standard 2', which illuminates the face with standard lighting at level 2, and `Cross-polarized', which reduces unwanted specular reflections on the skin.

We wanted to collect as many measurement points as possible from a face. As shown in Fig. \ref{fig:data_posid}, we collect TEWL and SH measurements at 37 anchor points on the face of each panelist. To standardize the collection of measurements, we provide each position with a specific ID, which plays a significant role when designing the model.

We followed a consistent protocol for aligning sticker placement and measurement collection on the face. Points 2 to 18 are located on the left side, while points 19 to 35 are on the right.  We collected data from five points around the eyebrows, with point 1 between the eyebrows, points 2 and 19 above them, and points 3 and 20 at their ends. Points 4, 5, 6, 21, 22, and 23 are consistently positioned below the eyes: points 4 and 21 near the nose, points 5 and 22 below the mid-eye, and points 6 and 23 aligned with the eyebrow's end. Points 7 and 24 are located in the crow's feet area, while points 36 and 37 are on the left and right eyelids, respectively. Points 8, 9, 10, 26, 27, and 28 form the second row below the eyes, with points 8 and 26 near the nasolabial junction. The third row on the cheeks consists of points 13, 14, 15, 30, 31, and 32, aligned with the nasolabial fold, with points 13 and 26 near the nasolabial and nose-lip junction. Points 4, 8, and 13 (or 21, 26, and 30 on the right) are always in line. Points 17 and 34 are positioned near the chin, within the smile zone, while points 18 and 35 are near the jawline. Lastly, points 11, 12, 16, 25, 29, and 33 form the final column near the ear, where TEWL measurements are typically favorable.

Through this collection process, we obtained a dataset that is consistent in the positioning of assessment points for SH/TEWL, comprehensive for full-face analysis, and diverse across different lighting conditions.

\subsection{Anchor Estimation and Skin Patches Cropping}\label{sec:centroids}
As shown in Fig. \ref{fig:centroids_model}, we proposed a pipeline to estimate measurement anchors, i.e., centroids for cropping skin patches from a facial image, based on the
detected facial landmarks. 
We use a robust head detection model, DAD-3Dheads~\cite{dad3dheads}, to identify 68 facial landmarks $\mathcal{P} = \{\boldsymbol{p}_j \in \mathbb{R}^2 \mid j=1,...,68\}$ on both sticker-labeled and non-sticker images. Next, we determine the centroids of the stickers in the sticker-labeled images with the aid of color segmentation. We then leverage the facial landmarks and sticker centroids of sticker-labeled images to train a PointNet~\cite{pointnet2017} model, denoted as $h$, which takes point data as input. Finally, the trained model can estimate the centroids of the points in non-sticker facial images by inputting the corresponding facial landmarks. Given a panelist's facial landmarks $\mathcal{P}_i$, the anchor estimation process is expressed as:
\begin{equation}
    \mathcal{C}_i = h(\mathcal{P}_i),
\end{equation}
where $\mathcal{C}_i = \{\boldsymbol{c}_j \in \mathbb{R}^2\ \mid j=1,...,M\}  $ represents the predicted centroid coordinates. \textcolor{black}{Here, $M$ is the number of anchor points per facial image and will be consistently used in the subsequent sections.}

\textcolor{black}{Once the centroids are identified, we crop skin patches from the non-sticker facial images. Let
$N$ be the number of samples in our dataset. We define the dataset as $\mathcal{S} = \{(\boldsymbol{I}_i, \mathcal{D}_i, \mathcal{C}_i, \mathcal{Y}_i)\}_{i=1}^N$, where 
$\boldsymbol{I}_i \in \mathbb{R}^{w \times h \times 3}$ denotes the facial images, 
$\mathcal{D}_i = \{d_j \in \mathbb{N} \mid j=1,...,M\}$ is the set of anchor IDs in the facial images, 
$\mathcal{C}_i = \{\boldsymbol{c}_{d_j} \in \mathbb{R}^2\ \mid d_j \in \mathcal{D}_i\}$ represents the set of corresponding anchors' coordinates and 
$\mathcal{Y}_i = \{y_{d_j} \in \mathbb{R}\ \mid d_j \in \mathcal{D}_i\}$ is the set of labels. The skin patch image is cropped as follows:
\begin{equation}
\boldsymbol{x}_{i,{d_j}} = \boldsymbol{I}_i[\boldsymbol{c}_{{d_j},0}-r:\boldsymbol{c}_{{d_j},0}+r, \quad \boldsymbol{c}_{{d_j},1}-r:\boldsymbol{c}_{{d_j},1}+r],    
\end{equation}
where $r$ is the semi-side length of the patches. }

\begin{figure*}[t!]
\centering
\includegraphics[width=0.75\textwidth]{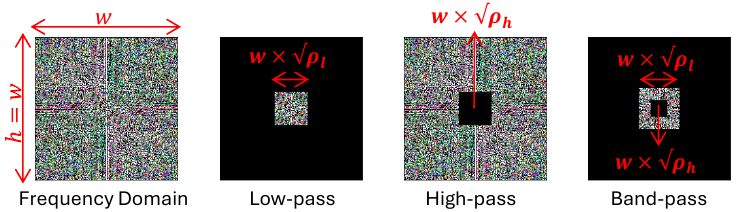}
\caption{\textcolor{black}{Three types of filters in the frequency domain. Given an image with dimension $w\times h$ (with $h=w$), the corresponding frequency domain image has the same size. Let $\rho_l$ and $\rho_h$ denote the proportions of frequency retained and removed, respectively. A low-pass filter retains only the central low-frequency region of size $w\times \sqrt{\rho_l}$, while a high-pass filter removes the central region of size $w\times \sqrt{\rho_h}$, keeping only the surrounding high-frequency components. A band-pass filter retains the frequency components that lie between $\rho_l$ and $\rho_h$.}
}\label{fig:filter}
\end{figure*}

\textcolor{black}{To ensure the skin patch dimensions used for SH/TEWL prediction are standardized regardless of the imaging modality, we use the real-world size of the stickers as a consistent reference for defining the patch crop dimension in different datasets. We determined the cropping dimensions based on the average sticker radius observed in each kind of sticker-label facial dataset.
In the end, we apply $r=70$ and $r=170$ to crop Selfie and VISIA images, respectively.
}


\subsection{Model Architecture}\label{sec:model_arch}
This section presents the proposed Skin-Prior Adaptive Vision Transformer (Skin-PAViT) for the SH/TEWL assessment. The core concept behind Skin-PAViT is to generate prompts for the Vision Transformer (ViT)~\cite{vit} by leveraging skin patches, along with the associated highlighted textures and positional information. 


As shown in Fig. \ref{fig:pipeline}\textbf{d}, we designed a straightforward Convolutional Neural Network to serve as a Prior Texture Module (PTM), capable of capturing local spatial contexts from input images to extract texture information. The module starts with a $3\times3$ convolution with 48 filters, followed by a pooling operation that halves the spatial dimensions. This pattern continues with progressively deeper convolutional layers (96, 192, 384, and 768 filters) interspersed with pooling layers that repeatedly reduce the spatial size, eventually down to $7 \times 7 \times D'$, where $D'$ represents the number of feature channels. 

\textcolor{black}{Our Texture Adapters (TAs) are designed to generate prompts to prepend the input sequence of each Transformer layer. The feature map of shape $7 \times 7 \times D'$ is flattened by combining the spatial dimensions ($7 \times 7$) into a single dimension, resulting in a new shape of $49 \times D'$. Each adapter consists of a simple Multi-Layer Perceptron (MLP) layer, adjusting the feature shape from $49 \times D'$ to $49 \times D$, where $D$ is the desired feature dimension of ViT's input token. These TAs combine with PTM to form a Texture Adaptive Module (TAM).}

To emphasize our model's capability of capturing skin texture, we used a band-pass filter as our texture highlighting operation, which retains a certain range of spatial frequency components, effectively preserving the structure and outline of skin texture, such as wrinkles and pores. Given an input image, $\boldsymbol{x}\in \mathbb{R}^{w \times h \times 3}$, we first convert the image into the frequency domain by Fourier Transform, $\mathtt{fft}$, and then shift the low frequencies to the center using the $\mathtt{fftshift}$ operator. This process is expressed as $\boldsymbol{f} = \mathtt{fftshift}(\mathtt{fft}(\boldsymbol{x}))$. We then generate a binary filtering mask $\textbf{M}\in\{0,1\}^{w\times h}$ to filter out the unwanted frequency component. The mask $\textbf{M}$ is defined as:
\begin{equation}
\textbf{M}=\textbf{M}_{\rho_l}\circ \textbf{M}_{\rho_h}
\end{equation}
where $\textbf{M}_{\rho_l}$ and $\textbf{M}_{\rho_h}$ are the masks for low-pass and high-pass filtering, respectively, with a corresponding $\rho$ representing the percentage of frequency components retained ($\rho_l$) or removed ($\rho_h$). Take note that $\rho_l$ is always greater than $\rho_h$ for band-pass filtering. 
\textcolor{black}{Given $u\in\{0,1,…,w-1\}$ and $v\in\{0,1,…,h-1\}$ denote the horizontal and vertical frequency indices in the frequency domain representation,} the masks are computed as:
\begin{equation}
    \textbf{M}_{\rho_l}(u,v) = \mathtt{mask}(\rho_l, u, v)    
\end{equation}
\begin{equation}
    \textbf{M}_{\rho_h}(u,v) = 1 - \mathtt{mask}(\rho_h, u, v)
\end{equation}
where $\mathtt{mask}$ is expressed as
\begin{equation}
\mathtt{mask}(\rho, u, v) =
\begin{cases} 
1, & \text{if } \frac{1-\sqrt{\rho}}{2}\cdot w \leq u \le \frac{1+\sqrt{\rho}}{2}\cdot w, \\
   & \text{and} \quad \frac{1-\sqrt{\rho}}{2}\cdot h \leq v \le \frac{1+\sqrt{\rho}}{2}\cdot h \\
0, & \text{otherwise}
\end{cases}
\end{equation}

Finally, the enhanced texture is obtained by applying the inverse frequency shifting operation $\mathtt{ifftshift}$ and inverse Fourier transform $\mathtt{ifft}$:
\begin{equation}
\boldsymbol{t} = \mathtt{ifft} ( \mathtt{ifftshift}(\boldsymbol{f} \times \textbf{M}) )
.
\label{eq:ifft} 
\end{equation}
A visual representation of the filtered frequency component, i.e., $\boldsymbol{f}\times\textbf{M}$, is presented in Fig. \ref{fig:filter}. The texture $\boldsymbol{t}$ is then concatenated with the input image and passed to the PTM.

Additionally, we introduce Position Adapters (PAs), each of which is also built using a simple MLP. \textcolor{black}{The PAs take anchor IDs, represented as one-hot encoded vectors, as input. Then, the output of PAs is concatenated with the output of TAs, resulting in 50 prompt tokens to prepend with the input sequences of each layer. 
Incorporating PAs enables each Transformer layer to be informed about the corresponding skin patch location.} As introduced in Section \ref{sec:data_collect_method}, the distribution of individual facial regions is different. Providing the model with the positional information of each skin patch is beneficial to enhancing contextual understanding.

\subsection{Cross-Lighting Generalization}\label{sec:lighting_aug}
Our study also aims to enable skin assessment via smartphone images in various environments, ensuring accessibility anytime and anywhere. To avoid overfitting to specific conditions, it's important to introduce diversity in the skin input patches. Since manually collecting facial images is costly and time-consuming, we incorporate four lighting augmentation operations to adjust the color saturation, contrast, brightness, and sharpness of the skin images. 
\textcolor{black}{The operations can be achieved by the blending of the degraded image and the original image. 
In the color saturation operation, the degraded image is a grayscale version of the original. For the contrast operation, it is a constant image set to the mean pixel value of the original. In the brightness operation, the degraded image is a completely black image (a zero-constant image). For the sharpness operation, it is a box-blurred version of the original.
The blending operation can be formulated as follows:
\begin{equation}
    \boldsymbol{x}_{adj} = \mathtt{clip} ((1-m) \times \boldsymbol{x}_{deg} + m \times \boldsymbol{x}_{ori},\hspace{0.25em}0,\hspace{0.25em}255),
\end{equation}
where $\boldsymbol{x}_{adj}$, $\boldsymbol{x}_{deg}$ and $\boldsymbol{x}_{ori}$ represent the augmented image, degraded image, and input image, respectively. The $\mathtt{clip}$ function ensures that the values of the augmented image remain within the range of $[0,255]$. Here, $m\in[0,2]$ is a magnitude parameter controlling the intensity of the adjustment: when $m=0$, the result is the degraded image; when $m=1$, it is the original images; when $m=2$, an enhanced image is produced. Examples of these lighting augmentation operations at various scales are provided in Appendix \ref{apx:light}. During training, one of the operations is chosen uniformly at random and applied to the image with a randomly selected magnitude from a uniform distribution. 
These lighting augmentation operations can be implemented using the ImageEnhance module from the Python Imaging Library Pillow~\cite{pil} in the training code.
}

\subsection{Objective Function}\label{sec:sym_con}
The SH/TEWL value distribution varies across different facial regions due to differences in skin properties, such as skin thickness. However, symmetric regions of the face tend to share similar distributions. To capture this in the feature space, we use contrastive learning to bring the latent features derived from symmetric skin patches of the same panelist closer to each other. Let the set of latent features be denoted as $Z$. Following ~\cite{supcon}, given a latent feature of a skin patch, $\boldsymbol{z_i}$, and its corresponding symmetric skin patch, $\boldsymbol{z_{i'}}$, where $\boldsymbol{z_i}, \boldsymbol{z_{i'}} \in Z$, the contrastive loss is formulated as:
\begin{equation}
\ell_{con}(\boldsymbol{z_i},\boldsymbol{z_{i'}}) = -\log \frac{\exp(\text{sim}(\boldsymbol{z_i},\boldsymbol{z_{i'}})/\tau)}{\sum_{\boldsymbol{z_k}\in Z, k \neq i}  \exp(\text{sim}(\boldsymbol{z_i},\boldsymbol{z_k})/\tau)}
,
\label{eq:con_loss} 
\end{equation}
where $\tau$ is the temperature parameter.


Similar to most of the regression tasks, we used Mean Squared Error(MSE) as the loss function for our model training. Combining with the proposed skin symmetric-based contrastive loss, the objective function of our model can be denoted as:
\begin{equation}
L_{total} = \frac{1}{N}\sum_{i=1}^{N}[L_{con}(\boldsymbol{z_i},\boldsymbol{z_{i'}}) + L_{MSE}(y_i, \hat{y_i})]
,
\label{eq:obj_func} 
\end{equation}
where $N$ is the total number of samples, and $y$ and $\hat{y}$ denote the actual and predicted label.

\subsection{Data augmentation}\label{sec:DA}
We applied data augmentation to the training data to reduce overfitting and improve generalization. Utilizing data augmentation in our task effectively alleviates the problem of imbalanced regression. The augmentation techniques used in our solution included common image transformations such as horizontal and vertical flipping, rotation, random erasing, and random cropping.
During training, each input image is applied with a randomly selected data augmentation operation, which has a probability of 0.5. 




\subsection{Heatmap Visualization}\label{sec:heatmap}
After predicting TEWL and SH values, we generate a heatmap based on these predictions and their corresponding centroids. Regions without data are interpolated linearly. We represent good skin conditions, i.e., well-hydrated or strong skin barrier function, using blue colors, and poor conditions using red colors. \textcolor{black}{The unit for TEWL is $g\cdot m^{-2}\cdot h^{-1}$ and that for SH is AU. For brevity, we omit the units in the following descriptions. We utilize Seismic colormaps for the TEWL heatmap, ranging from 0 to 30, with 15 set as the midpoint, and inverted Seismic colormaps for the SH heatmap, ranging from 0 to 90, with 45 as the midpoint. By doing so, SH values below 40 are shown in red, indicating dry skin. Values within the normal range of 40–50 appear in white, and those above 50 are depicted in blue, associated with well-hydrated skin. Similarly, TEWL values below 15 are visualized in blue, indicating good moisture retention, while values exceeding 15 are shown in red, suggesting compromised barrier function or very dry skin.}

\begin{figure}[t!]
\centering
\includegraphics[width=0.5\textwidth]{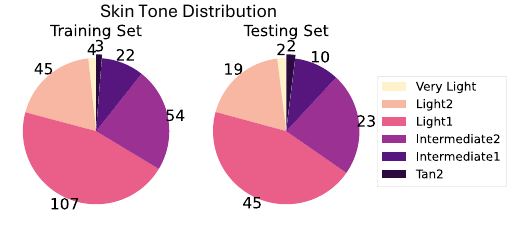}
\caption{We split the dataset into training and testing sets based on skin tone distribution to ensure both sets have similar skin tone representation.}\label{fig:data_split}
\end{figure}

\begin{figure}[t!]
\centering
\includegraphics[width=.5\textwidth]{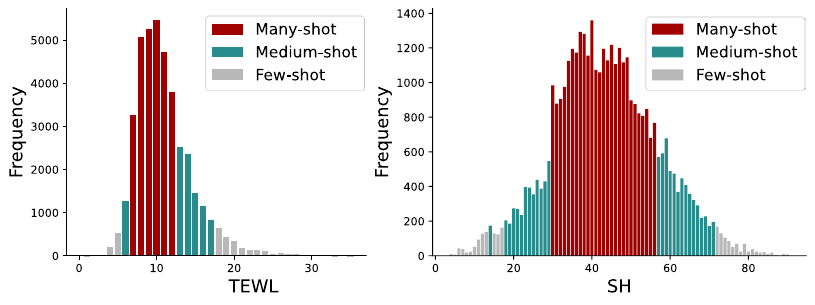}
\caption{Division of samples into diverse shot subgroups based on distribution frequency.}\label{fig:results_shot}
\end{figure}

\begin{figure*}[t!]
\centering
\includegraphics[width=0.75\textwidth]{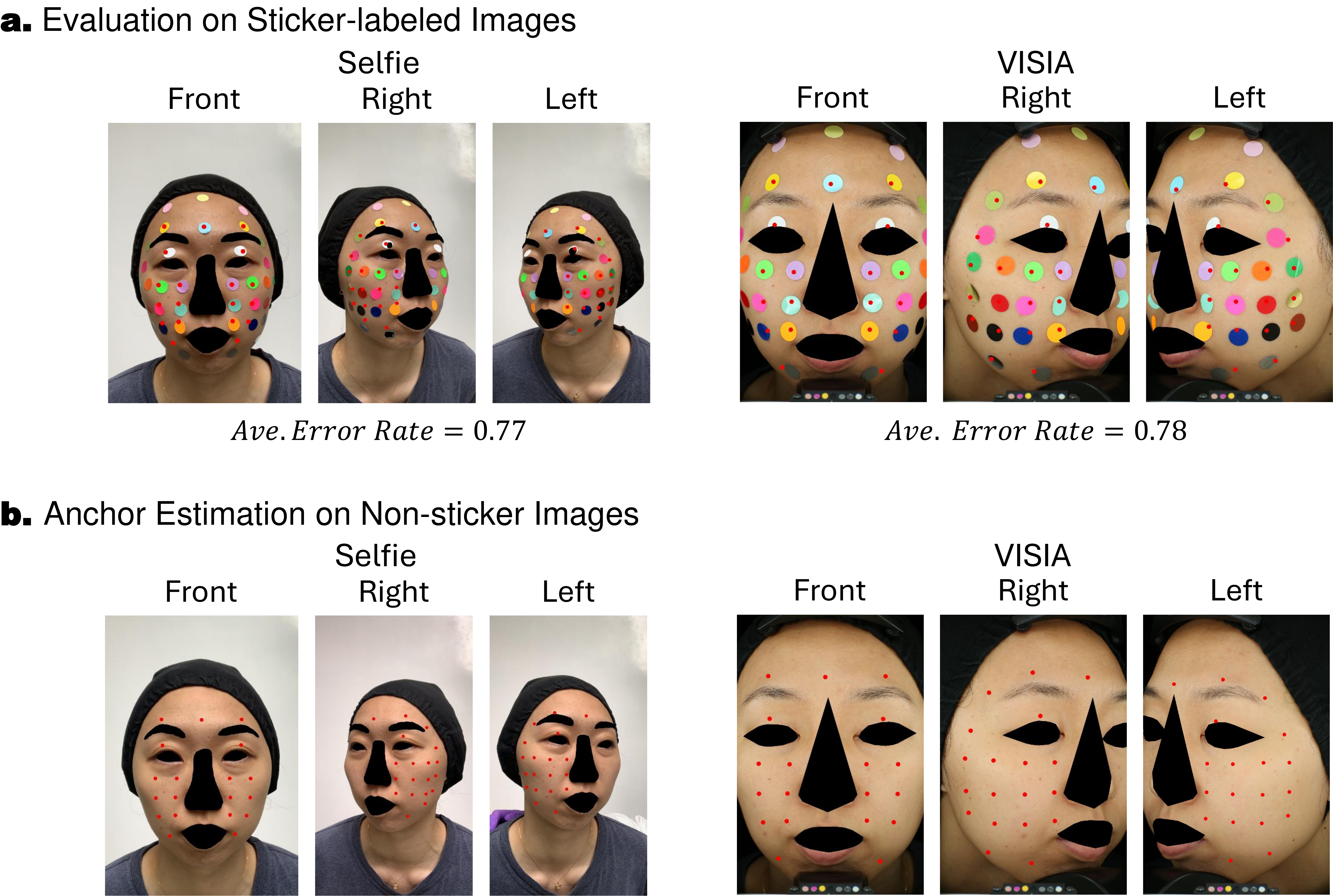}
\caption{\textbf{Anchor Estimation Result.}\textsuperscript{1} \textbf{a}, Visual examples of anchor estimation and average error rate performance for each facial angle of Selfie and VISIA datasets. The Error Rate represents the ratio of the error between the sticker centroid and the predicted centroid relative to the sticker's radius. \textbf{b}, Visual examples of anchors estimated by the PointNet on non-sticker images.}\label{fig:centroids_result}
\end{figure*}

\section{Result}\label{sec2}
\subsection{Dataset}\label{sec:data_collect_analysis}





\textbf{Data Splitting}. We collected facial images from 336 Chinese panelists. We then computed each panelist's ITA$^\circ$~\cite{ita} value, representing their skin tone level. According to its distribution, we divided the panelists into a training set of 235 and a test set of 101. Therefore, both training and testing sets maintain a similar skin tone distribution, as shown in Fig. \ref{fig:data_split}. 
As mentioned, the Selfie and VISIA images were captured from three facial angles under various lighting conditions. Hence, each panelist would have nine Selfie and six VISIA facial images. We can crop out 19 skin patches from each facial image, yielding that the Selfie dataset contains 40,185 skin patches for training and 17,271 for testing, and the VISIA dataset contains 26,790 and 11,514 skin patches for training and testing, respectively. 

\textcolor{black}{As mentioned in Section~\ref{sec:centroids},} for the Selfie dataset, the observed average sticker radius is $70$ pixels, so we cropped the skin patches from images at a resolution of $140\times140$ pixels. Similarly, for VISIA images, each patch was cropped at a resolution of  $340\times340$ pixels.

\textbf{Data Distribution Analysis}. Fig. \ref{fig:data_dist} illustrates the TEWL and SH value distributions of the training skin patch images from different datasets, where both value distributions exhibit distinct peaks. 
Based on the frequency of values in the distribution range, we divide the data into three disjoint groups: \textbf{many}, \textbf{medium}, and \textbf{few-shot} samples. Many-shot samples are those with a frequency greater than or equal to half of the maximum frequency in the training data distribution. Medium-shot samples have frequencies between one-quarter and one-half of the maximum, while few-shot samples have frequencies below one-quarter. Fig. \ref{fig:results_shot} visually represents this breakdown.

\textbf{Data Distribution Analysis on Individual Position}. After analyzing the data distribution for all skin patch images, we further examined the distribution for individual facial regions. We present a heatmap showing the average SH/TEWL values of individual facial regions across the dataset in Fig. \ref{fig:data_ave_heatmap}. From the figure, we observe that the heatmap intensity, which represents the values, is different across the face. For example, the average TEWL value tends to be higher on the eyelids and around the nose, while the average SH value is lower surrounding the eye region. These observations highlight the variation in value distributions across different facial regions.

\subsection{Evaluation Metric}
Two common regression metrics,  R-squared ($R^2$) score and Mean Absolute Error (MAE), are used to evaluate the performance of the SH/TEWL assessment model. The $R^2$ score represents the goodness of a model fit for the regression task, reflecting the strength of the relationship between the SH/TEWL assessor and the SH/TEWL values. The $R^2$ score could be negative when the model is arbitrarily worse. In addition, the MAE shows the average absolute differences between the actual and predicted values of a set of samples, clearly indicating the model's prediction accuracy on the set. 
Since our dataset is divided into three groups (many-shot, medium-shot, and few-shot samples), we also report the MAE for each group separately to evaluate the model's ability to handle imbalanced regression in this task.

\subsection{Training Details}
For the model training, we utilized an NVIDIA A5000 GPU to accelerate the process, leveraging its powerful computational capabilities for deep learning tasks. The implementation was done in PyTorch, which provided the flexibility needed to work with dynamic data. We scaled the labels to a range of 0-1 to align them with the model's output range, improving convergence during training. The learning process was guided by the Adam optimizer, with a learning rate set at 1e-5. To dynamically adjust the learning rate over the course of training, we employed the CosineAnnealingLR scheduler. The model was trained for 50 epochs with a batch size of 16, balancing gradient updates and memory efficiency to optimize performance.

\subsection{Performance of Anchor Estimation}\label{sec:result_centroids}
Using facial landmarks and corresponding sticker centroid coordinates as ground truth from the training set, we trained a PointNet model~\cite{pointnet2017}. The model was then evaluated on facial landmarks from the testing set’s sticker-labeled images.
The evaluation results across different datasets are shown in Fig. \ref{fig:centroids_result}\textbf{a}. Fig. \ref{fig:centroids_result}\textbf{b} provides examples of anchor estimation on non-sticker facial images.

To evaluate the model's performance, we compute the error rate, which represents the ratio of the error between the sticker centroid and the predicted anchor to the sticker's radius. \textcolor{black}{The error rate is defined as below:
\begin{equation}
Error Rate = \frac{Dist(\boldsymbol{c}, \boldsymbol{c'})}{r}
\label{eq:anchor} 
\end{equation}
where $\boldsymbol{c}$, $\boldsymbol{c'}$ and $r$ are the sticker centroid, the predicted anchor and the radius, respectively. The average error rates for Selfie and VISIA images are 0.77 and 0.78, respectively. Both error rates are acceptable, as values below 1 indicate that the anchor prediction error is within the radius of the stickers. As shown in a visualization of one panelist in Fig. \ref{fig:centroids_result}\textbf{a}, nearly all predicted anchors fall within the sticker boundaries.}

\textcolor{black}{
During data collection, we followed a standardized protocol for acquiring SH and TEWL measurements. However, since both measurement devices operate using point-based sampling, slight deviations between the actual measurement points and the sticker centroids are inevitable. Therefore, we did not aim for extremely precise centroid predictions in our anchor estimation model. Instead, we required that the model’s predicted points fall within the radius of the sticker, which is sufficient given the inherent variability in the data collection process.} 

\begin{figure}[t!]
\centering
\includegraphics[width=0.5\textwidth]{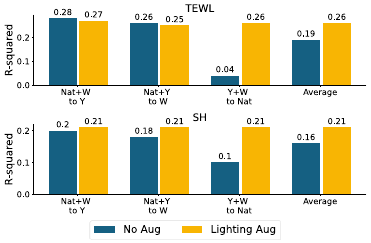}
\caption{The experimental results indicate that our introduced lighting augmentation effectively enhances generalization to other lighting domains. `Nat', `W', and `Y' indicate natural, white and yellow lighting, respectively.}\label{fig:results_light_aug}
\end{figure}

\subsection{Skin-Prior Adaptive Vision Transformer}\label{sec:result_overall}
As shown in Table \ref{tab:results}, our method outperforms traditional deep learning models in both TEWL and SH prediction across all datasets, as evidenced by the lowest MAE (All) and the highest $R^2$ score. Our proposed method significantly reduces the MAE for medium- and few-shot samples while maintaining competitive performance for many-shot samples. This demonstrates that our model is highly accurate and generalizable, outperforming traditional models such as ResNet, VGG-16, and other ViT-based architectures.

To provide an intuitive understanding of Skin-PAViT's effectiveness in addressing imbalance distribution, we compare its heatmaps to those from a fine-tuned ViT model in Fig. \ref{fig:results_heatmap} \textcolor{black}{and Fig. \ref{fig:results_heatmap_selfie}} for subjects whose ground truth heatmaps significantly differ from the average heatmap (Fig. \ref{fig:data_ave_heatmap}). These subjects have more regions where SH/TEWL values fall within the medium- and few-shot ranges. As illustrated, our model’s heatmaps align more closely with the ground truth, while the ViT model tends to resemble the average heatmap.

\subsection{Lighting Generalization}\label{sec:result_lighting}
We are also concerned about the lighting generalization to ensure our model's broader applicability. We proposed lighting augmentation to modify the appearance of skin patch images by adjusting their color, contrast, brightness, and sharpness. Examples of the different intensity transformations are provided in Appendix \ref{apx:light}. We evaluate the effectiveness of our lighting augmentation by leaving-one-lighting-out experiments on the Selfie dataset. As shown in Fig. \ref{fig:results_light_aug}, the experiments on `Nat+W to Y' and `Nat+Y to W' par with results without augmentation for TEWL prediction. However, other experiments indicate that incorporating lighting augmentation improves prediction performance. Overall, the average $R^2$ scores increased for both TEWL and SH prediction after including the lighting augmentation.

\begin{table*}[htbp]
\small
  \centering
  \caption{Performance changes with progressive configuration additions. The \textcolor{black}{`}+' Symbol denotes the incremental addition of model configurations. For instance, \textit{Config. A} consists of the frozen ViT with the trainable Texture Adaptive Module (TAM) and \textit{Config. B} retains the structure of the \textit{Config. A} but includes the frequency component as an additional input. Similarly, \textit{Config. C}, \textit{D}, and \textit{E} build on the preceding configurations, with `PAs', `Aug.' and `Sym.' representing Position Adapters, data augmentation and symmetric-based contrastive learning, respectively. Ultimately, our final models (\textit{Config. E}) achieve superior $R^2$ performance across all datasets, and the MAE for medium- and few-shot samples shows a decreasing trend as model configurations are incrementally added. $\downarrow$ indicates that lower values are better, while $\uparrow$ indicates that higher values are better.}
    \begin{tabular}{clrrrrr|rrrrr}
    \hline
       &    & \multicolumn{5}{c|}{TEWL} & \multicolumn{5}{c}{Skin Hydration} \\
    \hline
    \multicolumn{1}{l}{Dataset} & Methods & \multicolumn{1}{p{2.5em}}{MAE$\downarrow$ (All)} & \multicolumn{1}{p{2.5em}}{MAE$\downarrow$ (Many)} & \multicolumn{1}{p{2.5em}}{MAE$\downarrow$ (Med.)} & \multicolumn{1}{p{2.5em}}{MAE$\downarrow$ (Few)} & \multicolumn{1}{l|}{$R^2$$\uparrow$} & \multicolumn{1}{p{2.5em}}{MAE$\downarrow$ (All)} & \multicolumn{1}{p{2.5em}}{MAE$\downarrow$ (Many)} & \multicolumn{1}{p{2.5em}}{MAE$\downarrow$ (Med.)} & \multicolumn{1}{p{2.5em}}{MAE$\downarrow$ (Few)} & \multicolumn{1}{l}{$R^2$$\uparrow$} \\
    \hline
    \noalign{\vskip +0.8ex}
    \multirow{7}[4]{*}{Selfie} & ViT-B  & 2.38 & 1.89 & 2.57 & 8.44 & 0.209 & 9.33 & 6.31 & 16.66 & 26.78 & 0.133 \\
        \cline{2-12}
        \noalign{\vskip -0.8ex}
        & \multicolumn{11}{c}{\textcolor{black}{\textbf{ViT-B (Frozen)}}} \\
        \noalign{\vskip -0.8ex}
        \cline{2-12}
       &  A: + TAM & 2.38 & 1.81 & 2.72 & 8.55 & 0.204 & 9.50 & 6.30 & 17.10 & 26.48 & 0.108 \\
       &  B: + Freq. & 2.41 & 1.96 & 2.47 & 8.47 & 0.210 & 9.51 & 6.44 & 16.84 & 26.11 & 0.109 \\
       &  C: + PAs & 2.32 & 1.90 & 2.17 & 8.17 & 0.253 & 9.00 & 6.19 & 15.38 & 24.54 & 0.190 \\
       &  D: + Aug. & 2.30 & 1.91 & \textbf{2.05} & 8.11 & 0.262 & \textbf{8.95} & \textbf{6.10} & 15.49 & \textbf{24.50} & 0.201 \\
       &  E: + Sym. (\textbf{Ours}) & \textbf{2.28} & \textbf{1.62} & 2.41 & \textbf{6.49} & \textbf{0.264} & 8.97 & 6.33 & \textbf{15.04} & 24.92 & \textbf{0.205} \\
    \hline
    \noalign{\vskip +0.8ex}
    \multirow{7}[4]{*}{VISIA} & ViT-B  & 2.19 & 1.80 & 2.42 & 7.99 & 0.318 & 9.21 & 6.54 & 14.92 & 23.81 & 0.187 \\
        \cline{2-12}
        \noalign{\vskip -0.8ex}
        & \multicolumn{11}{c}{\textcolor{black}{\textbf{ViT-B (Frozen)}}} \\
        \noalign{\vskip -0.8ex}
        \cline{2-12}
       &  A: + TAM & 2.21 & 1.82 & 2.28 & 7.89 & 0.308 & 9.27 & 6.42 & 15.60 & 24.33 & 0.185 \\
       &  B: + Freq. & 2.21 & 1.82 & 2.39 & 7.91 & 0.311 & 9.14 & \textbf{6.18} & 15.80 & 24.26 & 0.200 \\
       &  C: + PAs & 2.18 & \textbf{1.77} & 2.28 & 7.78 & 0.320 & 8.99 & 6.40 & 14.76 & 23.01 & 0.231 \\
       &  D: + Aug. & \textbf{2.16} & 1.80 & 2.10 & 7.75 & 0.340 & 8.84 & 6.24 & 14.55 & 22.97 & 0.256 \\
       &  E: + Sym. (\textbf{Ours}) & 2.19 & 1.86 & \textbf{1.99} & \textbf{7.67} & \textbf{0.348} & \textbf{8.83} & 6.45 & \textbf{14.10} & \textbf{22.25} & \textbf{0.263} \\
    \hline
    \end{tabular}%
  \label{tab:results_ablation}%
\end{table*}%

\subsection{Ablation Study on the Skin-PAViT Architecture}\label{sec:result_eff_arc}
Skin-PAViT comprises a frozen ViT pretrained with ImageNet, a Texture Adaptive Module (TAM), and Position Adapters (PAs). Skin-PAViT also includes a frequency filtering operation that emphasizes the skin texture of the input patches, which is then concatenated with the input images to serve as the model's input. Additionally, we also leveraged data augmentation and proposed symmetric-based contrastive learning in our solution. Hence, in this section, 
\textcolor{black}{
to systematically evaluate the contributions of each component, we conduct a series of ablation experiments with different configurations, defined as follows:
\begin{itemize}
    \item \textit{Config. A}: ViT + Texture Adaptive Module
    \item \textit{Config. B}: \textit{Config. A} + Frequency Filtering
    \item \textit{Config. C}: \textit{Config. B} + Position Adapters 
    \item \textit{Config. D}: \textit{Config. C} + Data Augmentaion
    \item \textit{Config. E}: \textit{Config. D} + Symmetric-based Contrastive Learning
\end{itemize}
}

\textbf{Texture Adaptive Module.} The trainable TAM consists of a Prior Texture Module (PTM), which CNN builds up to learn fine-grained local details, such as wrinkles and pores in skin images, and Texture Adapters (TAs), which prompt the output of PTM to the corresponding transformer layers. As shown in Table. \ref{tab:results_ablation}, this approach (\textit{Config. A}) achieves comparable results to ViT while utilizing fewer tunable parameters than fully fine-tuning the ViT.

\begin{figure}[t!]
\centering
\includegraphics[width=0.45\textwidth]{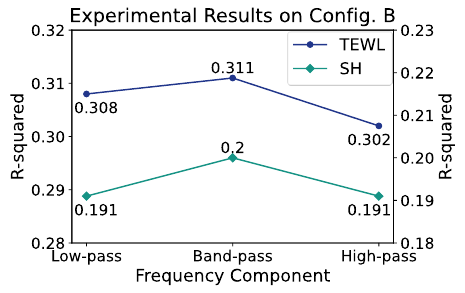}
\caption{The experimental results indicate that band-pass filtered textures are more beneficial for our models. }\label{fig:results_freq}
\end{figure}

\textbf{Highlighting Texture Information.} After configuring the initial prototype of Skin-PAViT (\textit{Config. A}), we tried to add on the operation for highlighting skin texture information and called it \textit{Config. B}. 
We enhance the representation of skin texture by filtering specific spatial frequency components. To determine the optimal frequency range for model training, we tested three different frequency ranges: low-pass at 5.76\%, band-pass from 0.36\% to 5.76\%, and high-pass above 0.36\% of the spatial frequency spectrum. 
Examples of the filtering outcomes are illustrated in Fig. \ref{fig:results_freq_exp}. 
Additional examples of the effects of various filtering ranges are provided in Appendix \ref{apx:freq}. 
The figure shows that excessive removal of frequency components results in blurred images when applying the low-pass filter and near-black images when using the high-pass filter. Since excessive removal of frequency components clearly fails to preserve skin texture, we only conduct experiments within an appropriate removal range.
We then combine the filtered result with the original skin image before passing them to the PTM. We experiment with the three filtering methods, and the corresponding results are depicted in Fig. \ref{fig:results_freq}. 
Incorporating the texture highlighted through band-pass filtering yields the highest $R^2$ scores for both SH and TEWL prediction. 
Eventually, we chose the band-pass filter as the operation for Skin-PAViT to effectively preserve the structure and outline of skin texture, such as wrinkles and pores. After incorporating the frequency-filtered images (\textit{Config. B}), we observe an increase in the $R^2$ values for both TEWL and SH predictions across both datasets, compared to \textit{Config. A}, as shown in Table \ref{tab:results_ablation}.

\textbf{Position Adapters.} On top of TAM and texture highlighting operation, we implemented Position Adapters (PAs) to prompt each transformer layer with skin patches positional data, labeling as \textit{Config. C}. As presented in Table \ref{tab:results_ablation}, when compared to the result without positional information (\textit{Config. B}), \textit{Config. C} demonstrates a significant improvement, as evidenced by an increase in $R^2$ for TEWL/SH predictions across both datasets.



After integrating all adapter modules into our model, we observe in Table \ref{tab:results_ablation} that the MAE for medium- and few-shot samples shows a downward trend from \textit{Config. A} to \textit{Config. C}. Building on this, we further discuss strategies to mitigate the impact of the imbalanced SH/TEWL distribution by reducing the MAE for medium- and few-shot samples.

\textbf{Data Augmentation.} In addition to lighting augmentation, we add on the common data augmentation techniques, including horizontal and vertical flipping, rotation, random erasing, and random cropping during the training process, to alleviate the imbalance issue and reduce overfitting to the many-shot samples, identifying as \textit{Config. D}.
From Table \ref{tab:results_ablation}, we observe a notable improvement in the MAE for almost all medium- and few-shot samples after applying data augmentation compared to \textit{Config. C}, which does not train with data augmentation. Furthermore, all $R^2$ values show a remarkable increase, indicating an overall enhancement in model performance.


\begin{figure}[t!]
\centering
\includegraphics[width=0.45\textwidth]{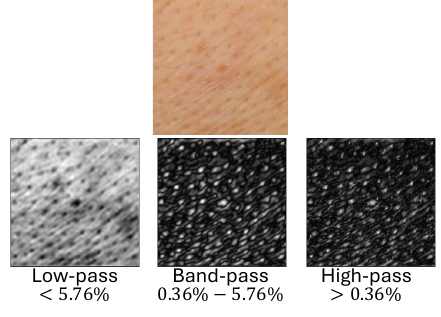}
\caption{Diverse Frequency Components Examples. Band-pass filtering effectively highlights skin texture compared to low-pass filtering while simultaneously removing noise in the texture through its high-pass component.}\label{fig:results_freq_exp}
\end{figure}

\textbf{Symmetric-based Contrastive Learning.} Furthermore, we considered the symmetrical properties of the face. As mentioned in the previous section, SH/TEWL value distributions vary across different facial regions, but symmetrical regions should exhibit similar value distributions. As shown in the average heatmap (Fig. \ref{fig:data_ave_heatmap}), the symmetric facial regions tend to be similar. To leverage this property, we implemented a facial symmetry-based contrastive learning technique, encouraging the latent representations of symmetric skin patches to be closely aligned. 
As presented in Table \ref{tab:results_ablation}, incorporating the technique into our training process (\textit{Config. E}) led to a reduction in MAE for TEWL prediction in the few-shot samples of the Selfie dataset and SH prediction in the medium-shot samples from the same dataset. In the VISIA dataset, both few-shot and medium-shot samples exhibited reduced MAE for predicting both TEWL and SH. Significantly, this final model configuration (\textit{Config. E}) of Skin-PAViT achieved superior performance, showing the greatest improvement in $R^2$ scores compared to all previous configurations.

\section{Conclusion and Future Work}

Our proposed method offers a novel solution to measure SH and TEWL across facial zones while overcoming the limitations of sensor-based methods. Our model can also make accurate predictions from a single view, though it was trained on three facial views to capture comprehensive features. This flexibility highlights the robustness of our method and its suitability for practical applications, such as diagnostic tools utilizing common capturing devices like smartphones. 

\textcolor{black}{
Our study presents a promising step toward AI-based remote facial skin assessment, though several directions remain for future exploration. While our goal was to develop a generalizable solution for home-use diagnosis, resource constraints limited our Selfie dataset, which only captured under three lighting conditions. Expanding this dataset to include more diverse and uncontrolled environments would improve the method's robustness and generalizability.
}

In our experiment, the predictive accuracy for Selfie images is notably lower than for VISIA images. This discrepancy can be attributed to several factors. VISIA images are taken under controlled lighting conditions, with consistent face size and orientation, whereas Selfie images are more variable and difficult to standardize.
\textcolor{black}{
These differences highlight the need to improve SH/TEWL estimation under less controlled conditions, which is essential for real-world deployment. } 

We also observed that SH prediction is less accurate than TEWL, consistent with prior findings~\cite {TDA}, which attributes this gap to SH’s weaker correlation with skin texture.
Our method emphasizes surface-level skin texture, which can be effectively highlighted through frequency filtering. However, it does not account for skin radiance and tone, which are more closely associated with SH~\cite{skinhydration, skinmap}. 
\textcolor{black}{
Therefore, future work could explore incorporating more non-RGB modalities (e.g., infrared) and leveraging advanced multimodal fusion techniques~\cite{fusionmamba} to enhance predictive performance.
}

In conclusion, our findings demonstrate the potential of deep learning to assess skin hydration and barrier function via facial images and emphasize the importance of artificial intelligence in revolutionizing skin care and treatment. \textcolor{black}{In future, we would also investigate artificial intelligence in other skin related topics, such as skin prototype segmentation \cite{Zheng2022}, skin product efficacy prediction \cite{Li2023}, simulating the recovering process of skin pore or acne~\cite{shuai}.}

\section{Data Availability}
\textcolor{black}{The facial images collected and used in this paper contain personally identifiable information (PII) of participants. To protect participant privacy, the dataset is not planned to be publicly available according to the PII data protection policy. Additionally, any participant images presented in this paper have been masked to remove PII. Nevertheless, we provide a detailed and transparent description of our data collection and processing procedures, allowing other researchers to reference and build upon our methodology.} 

\section{Acknowledgment}
This work was carried out at the Rapid-Rich Object Search (ROSE) Lab, School of Electrical \& Electronic Engineering, Nanyang Technological University (NTU), Singapore in collaboration with P\&G . The research is supported in part by the A*STAR under it’s A*STAR-P\&G Collaboration (Award H23HW10004). Any opinions, findings and conclusions or recommendations expressed in this material are those of the author(s) and do not reflect the views of the A*STAR.





\backmatter

\bibliographystyle{ieeetr}
\bibliography{sn-bibliography}

\begin{thebibliography}{10}

\bibitem{correlateshtewl}
E.~Caberlotto, C.~Cornillon, S.~Njikeu, M.~Monot, M.~Vicic, and F.~Flament, ``Synchronized in vivo measurements of skin hydration and trans-epidermal water loss. exploring their mutual influences,'' {\em International Journal of Cosmetic Science}, vol.~41, no.~5, pp.~437--442, 2019.

\bibitem{tewl}
M.~Akdeniz, S.~Gabriel, A.~Lichterfeld-Kottner, U.~Blume-Peytavi, and J.~Kottner, ``Transepidermal water loss in healthy adults: a systematic review and meta-analysis update,'' {\em British Journal of Dermatology}, vol.~179, no.~5, pp.~1049--1055, 2018.

\bibitem{skinmap}
R.~Voegeli, J.~Gierschendorf, B.~Summers, and A.~Rawlings, ``Facial skin mapping: from single point bio-instrumental evaluation to continuous visualization of skin hydration, barrier function, skin surface ph, and sebum in different ethnic skin types,'' {\em International journal of cosmetic science}, vol.~41, no.~5, pp.~411--424, 2019.

\bibitem{skinhydration}
S.~Verdier-S{\'e}vrain and F.~Bont{\'e}, ``Skin hydration: a review on its molecular mechanisms,'' {\em Journal of cosmetic dermatology}, vol.~6, no.~2, pp.~75--82, 2007.

\bibitem{poresizeRednessTewl}
K.~Miyamoto, Y.~Munakata, X.~Yan, G.~Tsuji, and M.~Furue, ``Enhanced fluctuations in facial pore size, redness, and tewl caused by mask usage are normalized by the application of a moisturizer,'' {\em Journal of Clinical Medicine}, vol.~11, no.~8, p.~2121, 2022.

\bibitem{TDA}
K.~Koseki, H.~Kawasaki, T.~Atsugi, M.~Nakanishi, M.~Mizuno, E.~Naru, T.~Ebihara, M.~Amagai, and E.~Kawakami, ``Assessment of skin barrier function using skin images with topological data analysis,'' {\em NPJ systems biology and applications}, vol.~6, no.~1, p.~40, 2020.

\bibitem{vit}
A.~Dosovitskiy, L.~Beyer, A.~Kolesnikov, D.~Weissenborn, X.~Zhai, T.~Unterthiner, M.~Dehghani, M.~Minderer, G.~Heigold, S.~Gelly, {\em et~al.}, ``An image is worth 16x16 words: Transformers for image recognition at scale,'' {\em arXiv preprint arXiv:2010.11929}, 2020.

\bibitem{evp}
W.~Liu, X.~Shen, C.-M. Pun, and X.~Cun, ``Explicit visual prompting for low-level structure segmentations,'' in {\em Proceedings of the IEEE/CVF Conference on Computer Vision and Pattern Recognition}, pp.~19434--19445, 2023.

\bibitem{vpt}
M.~Jia, L.~Tang, B.-C. Chen, C.~Cardie, S.~Belongie, B.~Hariharan, and S.-N. Lim, ``Visual prompt tuning,'' in {\em European Conference on Computer Vision}, pp.~709--727, Springer, 2022.

\bibitem{vitAdapter}
Z.~Chen, Y.~Duan, W.~Wang, J.~He, T.~Lu, J.~Dai, and Y.~Qiao, ``Vision transformer adapter for dense predictions,'' {\em arXiv preprint arXiv:2205.08534}, 2022.

\bibitem{convpass}
S.~Jie and Z.-H. Deng, ``Convolutional bypasses are better vision transformer adapters,'' {\em arXiv preprint arXiv:2207.07039}, 2022.

\bibitem{resnet}
K.~He, X.~Zhang, S.~Ren, and J.~Sun, ``Deep residual learning for image recognition,'' in {\em Proceedings of the IEEE conference on computer vision and pattern recognition}, pp.~770--778, 2016.

\bibitem{vgg}
K.~Simonyan and A.~Zisserman, ``Very deep convolutional networks for large-scale image recognition,'' {\em arXiv preprint arXiv:1409.1556}, 2014.

\bibitem{effnet}
M.~Tan and Q.~Le, ``Efficientnet: Rethinking model scaling for convolutional neural networks,'' in {\em International conference on machine learning}, pp.~6105--6114, PMLR, 2019.

\bibitem{convnext}
Z.~Liu, H.~Mao, C.-Y. Wu, C.~Feichtenhofer, T.~Darrell, and S.~Xie, ``A convnet for the 2020s,'' in {\em Proceedings of the IEEE/CVF conference on computer vision and pattern recognition}, pp.~11976--11986, 2022.

\bibitem{swin}
Z.~Liu, Y.~Lin, Y.~Cao, H.~Hu, Y.~Wei, Z.~Zhang, S.~Lin, and B.~Guo, ``Swin transformer: Hierarchical vision transformer using shifted windows,'' in {\em Proceedings of the IEEE/CVF international conference on computer vision}, pp.~10012--10022, 2021.

\bibitem{pvt}
W.~Wang, E.~Xie, X.~Li, D.-P. Fan, K.~Song, D.~Liang, T.~Lu, P.~Luo, and L.~Shao, ``Pyramid vision transformer: A versatile backbone for dense prediction without convolutions,'' in {\em Proceedings of the IEEE/CVF international conference on computer vision}, pp.~568--578, 2021.

\bibitem{corneo_metrics}
U.~Heinrich, U.~Koop, M.-C. Leneveu-Duchemin, K.~Osterrieder, S.~Bielfeldt, C.~Chkarnat, J.~Degwert, D.~H{\"a}ntschel, S.~Jaspers, H.-P. Nissen, {\em et~al.}, ``Multicentre comparison of skin hydration in terms of physical-, physiological-and product-dependent parameters by the capacitive method (corneometer cm 825),'' {\em International journal of cosmetic science}, vol.~25, no.~1-2, pp.~45--53, 2003.

\bibitem{vapo_metrics}
T.~Klotz, A.~Ibrahim, G.~Maddern, Y.~Caplash, and M.~Wagstaff, ``Devices measuring transepidermal water loss: A systematic review of measurement properties,'' {\em Skin Research and Technology}, vol.~28, no.~4, pp.~497--539, 2022.

\bibitem{transformer}
A.~Vaswani, N.~Shazeer, N.~Parmar, J.~Uszkoreit, L.~Jones, A.~N. Gomez, {\L}.~Kaiser, and I.~Polosukhin, ``Attention is all you need,'' {\em Advances in neural information processing systems}, vol.~30, 2017.

\bibitem{bert}
J.~Devlin, M.-W. Chang, K.~Lee, and K.~Toutanova, ``Bert: Pre-training of deep bidirectional transformers for language understanding,'' in {\em Proceedings of the 2019 conference of the North American chapter of the association for computational linguistics: human language technologies, volume 1 (long and short papers)}, pp.~4171--4186, 2019.

\bibitem{t5}
C.~Raffel, N.~Shazeer, A.~Roberts, K.~Lee, S.~Narang, M.~Matena, Y.~Zhou, W.~Li, and P.~J. Liu, ``Exploring the limits of transfer learning with a unified text-to-text transformer,'' {\em Journal of machine learning research}, vol.~21, no.~140, pp.~1--67, 2020.

\bibitem{gpt3}
T.~Brown, B.~Mann, N.~Ryder, M.~Subbiah, J.~D. Kaplan, P.~Dhariwal, A.~Neelakantan, P.~Shyam, G.~Sastry, A.~Askell, {\em et~al.}, ``Language models are few-shot learners,'' {\em Advances in neural information processing systems}, vol.~33, pp.~1877--1901, 2020.

\bibitem{llama}
H.~Touvron, T.~Lavril, G.~Izacard, X.~Martinet, M.-A. Lachaux, T.~Lacroix, B.~Rozi{\`e}re, N.~Goyal, E.~Hambro, F.~Azhar, {\em et~al.}, ``Llama: Open and efficient foundation language models,'' {\em arXiv preprint arXiv:2302.13971}, 2023.

\bibitem{mvitv2}
Y.~Li, C.-Y. Wu, H.~Fan, K.~Mangalam, B.~Xiong, J.~Malik, and C.~Feichtenhofer, ``Mvitv2: Improved multiscale vision transformers for classification and detection,'' in {\em Proceedings of the IEEE/CVF conference on computer vision and pattern recognition}, pp.~4804--4814, 2022.

\bibitem{imagenet}
O.~Russakovsky, J.~Deng, H.~Su, J.~Krause, S.~Satheesh, S.~Ma, Z.~Huang, A.~Karpathy, A.~Khosla, M.~Bernstein, {\em et~al.}, ``Imagenet large scale visual recognition challenge,'' {\em International journal of computer vision}, vol.~115, pp.~211--252, 2015.

\bibitem{nlp-adapter}
N.~Houlsby, A.~Giurgiu, S.~Jastrzebski, B.~Morrone, Q.~De~Laroussilhe, A.~Gesmundo, M.~Attariyan, and S.~Gelly, ``Parameter-efficient transfer learning for nlp,'' in {\em International conference on machine learning}, pp.~2790--2799, PMLR, 2019.

\bibitem{nlp-parallel_adapter}
J.~He, C.~Zhou, X.~Ma, T.~Berg-Kirkpatrick, and G.~Neubig, ``Towards a unified view of parameter-efficient transfer learning,'' {\em arXiv preprint arXiv:2110.04366}, 2021.

\bibitem{nlp-lora}
E.~J. Hu, Y.~Shen, P.~Wallis, Z.~Allen-Zhu, Y.~Li, S.~Wang, L.~Wang, W.~Chen, {\em et~al.}, ``Lora: Low-rank adaptation of large language models.,'' {\em ICLR}, vol.~1, no.~2, p.~3, 2022.

\bibitem{nlp-prefix}
X.~L. Li and P.~Liang, ``Prefix-tuning: Optimizing continuous prompts for generation,'' {\em arXiv preprint arXiv:2101.00190}, 2021.

\bibitem{nlp-prompt_prepend}
B.~Lester, R.~Al-Rfou, and N.~Constant, ``The power of scale for parameter-efficient prompt tuning,'' {\em arXiv preprint arXiv:2104.08691}, 2021.

\bibitem{nlp-p-tunning}
X.~Liu, Y.~Zheng, Z.~Du, M.~Ding, Y.~Qian, Z.~Yang, and J.~Tang, ``Gpt understands, too,'' {\em AI Open}, vol.~5, pp.~208--215, 2024.

\bibitem{nlp-p-tunning2}
X.~Liu, K.~Ji, Y.~Fu, W.~L. Tam, Z.~Du, Z.~Yang, and J.~Tang, ``P-tuning v2: Prompt tuning can be comparable to fine-tuning universally across scales and tasks,'' {\em arXiv preprint arXiv:2110.07602}, 2021.

\bibitem{nlp-llama-adapter}
R.~Zhang, J.~Han, C.~Liu, A.~Zhou, P.~Lu, Y.~Qiao, H.~Li, and P.~Gao, ``Llama-adapter: Efficient fine-tuning of large language models with zero-initialized attention,'' in {\em The Twelfth International Conference on Learning Representations}, 2024.

\bibitem{C1}
A.~Luo, R.~Cai, C.~Kong, Y.~Ju, X.~Kang, J.~Huang, and A.~C.~K. Life, ``Forgery-aware adaptive learning with vision transformer for generalized face forgery detection,'' {\em IEEE Transactions on Circuits and Systems for Video Technology}, 2024.

\bibitem{C2}
R.~Cai, C.~Soh, Z.~Yu, H.~Li, W.~Yang, and A.~C. Kot, ``Towards data-centric face anti-spoofing: Improving cross-domain generalization via physics-based data synthesis,'' {\em International Journal of Computer Vision}, pp.~1--22, 2024.

\bibitem{C3}
X.~Lin, S.~Wang, R.~Cai, Y.~Liu, Y.~Fu, W.~Tang, Z.~Yu, and A.~Kot, ``Suppress and rebalance: Towards generalized multi-modal face anti-spoofing,'' in {\em Proceedings of the IEEE/CVF Conference on Computer Vision and Pattern Recognition}, pp.~211--221, 2024.

\bibitem{C4}
Z.~Yu, R.~Cai, Y.~Cui, A.~Liu, and C.~Chen, ``Visual prompt flexible-modal face anti-spoofing,'' {\em IEEE Transactions on Dependable and Secure Computing}, 2024.

\bibitem{C5}
R.~Cai, Z.~Yu, C.~Kong, H.~Li, C.~Chen, Y.~Hu, and A.~C. Kot, ``S-adapter: Generalizing vision transformer for face anti-spoofing with statistical tokens,'' {\em IEEE Transactions on Information Forensics and Security}, 2024.

\bibitem{C6}
R.~Cai, Y.~Cui, Z.~Li, Z.~Yu, H.~Li, Y.~Hu, and A.~Kot, ``Rehearsal-free domain continual face anti-spoofing: Generalize more and forget less,'' in {\em Proceedings of the IEEE/CVF International Conference on Computer Vision}, pp.~8037--8048, 2023.

\bibitem{torgo2013smote}
L.~Torgo, R.~P. Ribeiro, B.~Pfahringer, and P.~Branco, ``Smote for regression,'' in {\em Portuguese conference on artificial intelligence}, pp.~378--389, Springer, 2013.

\bibitem{branco2017smogn}
P.~Branco, L.~Torgo, and R.~P. Ribeiro, ``Smogn: a pre-processing approach for imbalanced regression,'' in {\em First international workshop on learning with imbalanced domains: Theory and applications}, pp.~36--50, PMLR, 2017.

\bibitem{branco2018rebagg}
P.~Branco, L.~Torgo, and R.~P. Ribeiro, ``Rebagg: Resampled bagging for imbalanced regression,'' in {\em Second International Workshop on Learning with Imbalanced Domains: Theory and Applications}, pp.~67--81, PMLR, 2018.

\bibitem{steininger2021density}
M.~Steininger, K.~Kobs, P.~Davidson, A.~Krause, and A.~Hotho, ``Density-based weighting for imbalanced regression,'' {\em Machine Learning}, vol.~110, pp.~2187--2211, 2021.

\bibitem{yang2021delving}
Y.~Yang, K.~Zha, Y.~Chen, H.~Wang, and D.~Katabi, ``Delving into deep imbalanced regression,'' in {\em International conference on machine learning}, pp.~11842--11851, PMLR, 2021.

\bibitem{ren2022balanced}
J.~Ren, M.~Zhang, C.~Yu, and Z.~Liu, ``Balanced mse for imbalanced visual regression,'' in {\em Proceedings of the IEEE/CVF Conference on Computer Vision and Pattern Recognition}, pp.~7926--7935, 2022.

\bibitem{gong2022ranksim}
Y.~Gong, G.~Mori, and F.~Tung, ``Ranksim: Ranking similarity regularization for deep imbalanced regression,'' {\em arXiv preprint arXiv:2205.15236}, 2022.

\bibitem{wang2023variational}
Z.~Wang and H.~Wang, ``Variational imbalanced regression: Fair uncertainty quantification via probabilistic smoothing,'' {\em Advances in Neural Information Processing Systems}, vol.~36, pp.~30429--30452, 2023.

\bibitem{dad3dheads}
T.~Martyniuk, O.~Kupyn, Y.~Kurlyak, I.~Krashenyi, J.~Matas, and V.~Sharmanska, ``Dad-3dheads: A large-scale dense, accurate and diverse dataset for 3d head alignment from a single image,'' in {\em Proceedings of the IEEE/CVF Conference on computer vision and pattern recognition}, pp.~20942--20952, 2022.

\bibitem{pointnet2017}
C.~R. Qi, H.~Su, K.~Mo, and L.~J. Guibas, ``Pointnet: Deep learning on point sets for 3d classification and segmentation,'' in {\em Proceedings of the IEEE conference on computer vision and pattern recognition}, pp.~652--660, 2017.

\bibitem{pil}
A.~Clark and Contributors, ``Pillow (python imaging library).'' \url{https://python-pillow.org}, 2010.
\newblock Version 9.2.0.

\bibitem{supcon}
P.~Khosla, P.~Teterwak, C.~Wang, A.~Sarna, Y.~Tian, P.~Isola, A.~Maschinot, C.~Liu, and D.~Krishnan, ``Supervised contrastive learning,'' {\em Advances in neural information processing systems}, vol.~33, pp.~18661--18673, 2020.

\bibitem{ita}
Y.~Wu, T.~Tanaka, and M.~Akimoto, ``Utilization of individual typology angle (ita) and hue angle in the measurement of skin color on images,'' {\em bioimages}, vol.~28, pp.~1--8, 2020.

\bibitem{fusionmamba}
X.~Xie, Y.~Cui, T.~Tan, X.~Zheng, and Z.~Yu, ``Fusionmamba: Dynamic feature enhancement for multimodal image fusion with mamba,'' {\em Visual Intelligence}, vol.~2, no.~1, p.~37, 2024.

\bibitem{Zheng2022}
Q.~Zheng, A.~Purwar, H.~Zhao, G.~L. Lim, L.~Li, D.~Behera, Q.~Wang, M.~Tan, R.~Cai, J.~Werner, D.~Sng, M.~van Steensel, W.~Lin, and A.~C. Kot, ``Automatic facial skin feature detection for everyone,'' in {\em Proc. IS\&T Int’l. Symp. on Electronic Imaging: Imaging and Multimedia Analytics at the Edge}, pp.~300--1--300--6, 2022.

\bibitem{Li2023}
L.~Li, B.~Dissanayake, T.~Omotezako, Y.~Zhong, Q.~Zhang, R.~Cai, Q.~Zheng, D.~Sng, W.~Lin, Y.~Wang, and A.~C. Kot, ``Evaluating the efficacy of skincare product: A realistic short-term facial pore simulation,'' {\em Electronic Imaging}, pp.~276--1--276--6, 2023.

\bibitem{shuai}
C.~Shuai, R.~Cai, B.~Dissanayake, A.~Newman, D.~Guan, D.~Sng, L.~Li, and A.~Kot, ``Controllable and gradual facial blemishes retouching via physics-based modelling,'' in {\em 2024 IEEE International Conference on Multimedia and Expo (ICME)}, pp.~1--6, 2024.

\end{thebibliography}

\section*{Biography}

\includegraphics[width=1in,height=1.25in,clip,keepaspectratio]{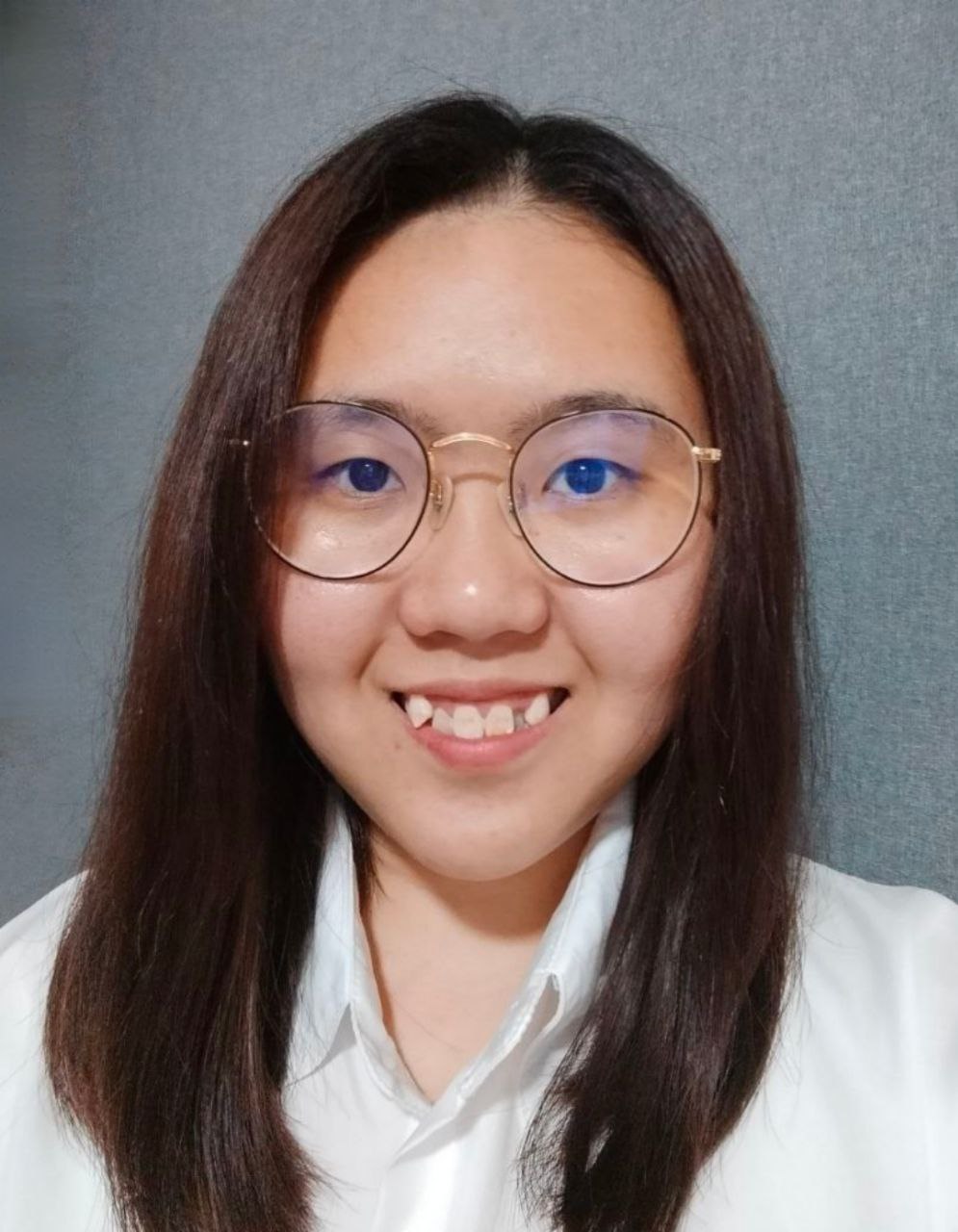}\\
\noindent
\textbf{Cecelia Soh} received her B.Eng. degree in Computer Engineering from Nanyang Technological University, Singapore, in June 2023. She worked as a Project Officer at the Rapid-
Rich Object Search (ROSE) Lab from July 2023 to December 2024. She is currently pursuing a Master of Computing in Artificial Intelligence at the National University of Singapore. Her research interests include deep learning and computer vision.\\
E-mail: ceceliasoh1030@gmail.com\\
ORCID iD: 0009-0004-6320-8164

\hfill \break

\includegraphics[width=1in,height=1.25in,clip,keepaspectratio]{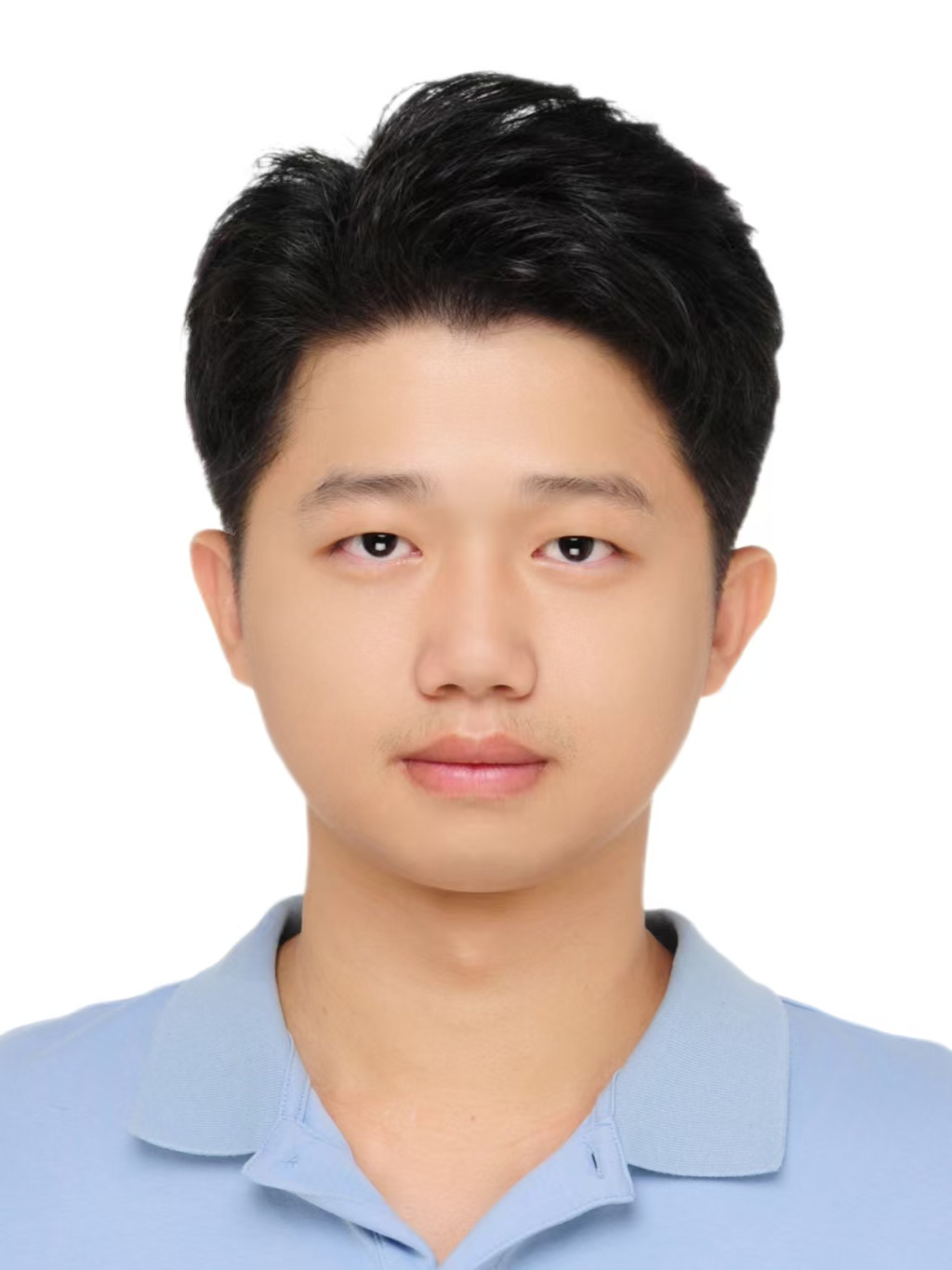}\\
\noindent
\textbf{Rizhao Cai} received his B.Eng degree in Electronic Information Engineering from  Shenzhen University, China in  2018, and Ph.D. degree from School of Electrical and Electronic Engineering, Nanyang  Technological  University, Singapore, in 2024.  Now, he is a research fellow in the Rapid-Rich Object Search (ROSE) Lab and NTU-PKU  Joint  Research  Institute, leading computer vision projects collaborated with industrial partners. His research interests include computer vision and biometric/AI security.\\
E-mail: rizhao001@e.ntu.edu.sg\\
ORCID iD: 0000-0002-7114-8462

\hfill \break

\includegraphics[width=1in,height=1.25in,clip,keepaspectratio]{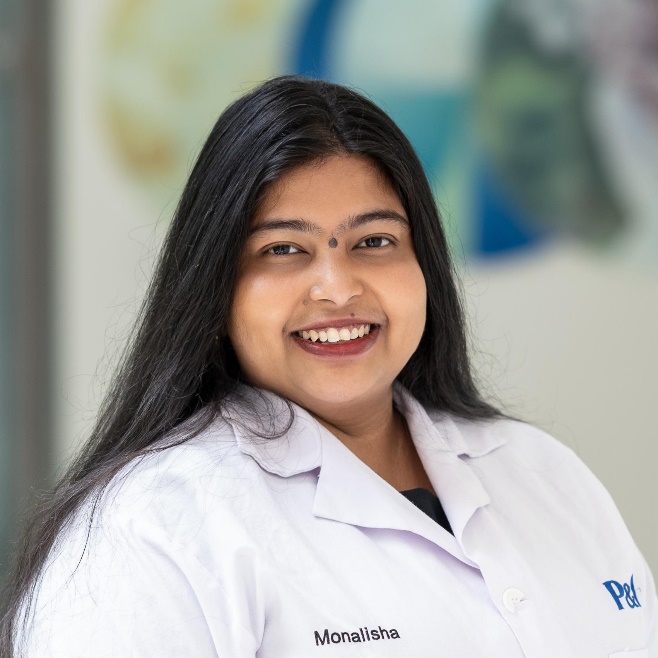}\\
\noindent
\textbf{Monalisha Paul}\\
\textbf{Scientist, Beauty Revealed Measurement Organization }\\
\textbf{Procter \& Gamble, Singapore}\\
Monalisha Paul holds a B. Tech in Cosmetic Technology from India and has over 20 years of experience in the skincare research industry, with more than 15 years at Procter \& Gamble. She specializes in formulation, method development, and claim support in skincare. Monalisha is passionate about leveraging technology to enhance consumer experiences in skincare, focusing on innovative solutions for measuring and tracking skin health anytime, anywhere.

In addition to her professional endeavours, she enjoys exploring advancements in cosmetic technology and their applications in improving consumer products.\\
E-MAIL: paul.m.9@pg.com

\hfill \break

\includegraphics[width=1in,height=1.25in,clip,keepaspectratio]{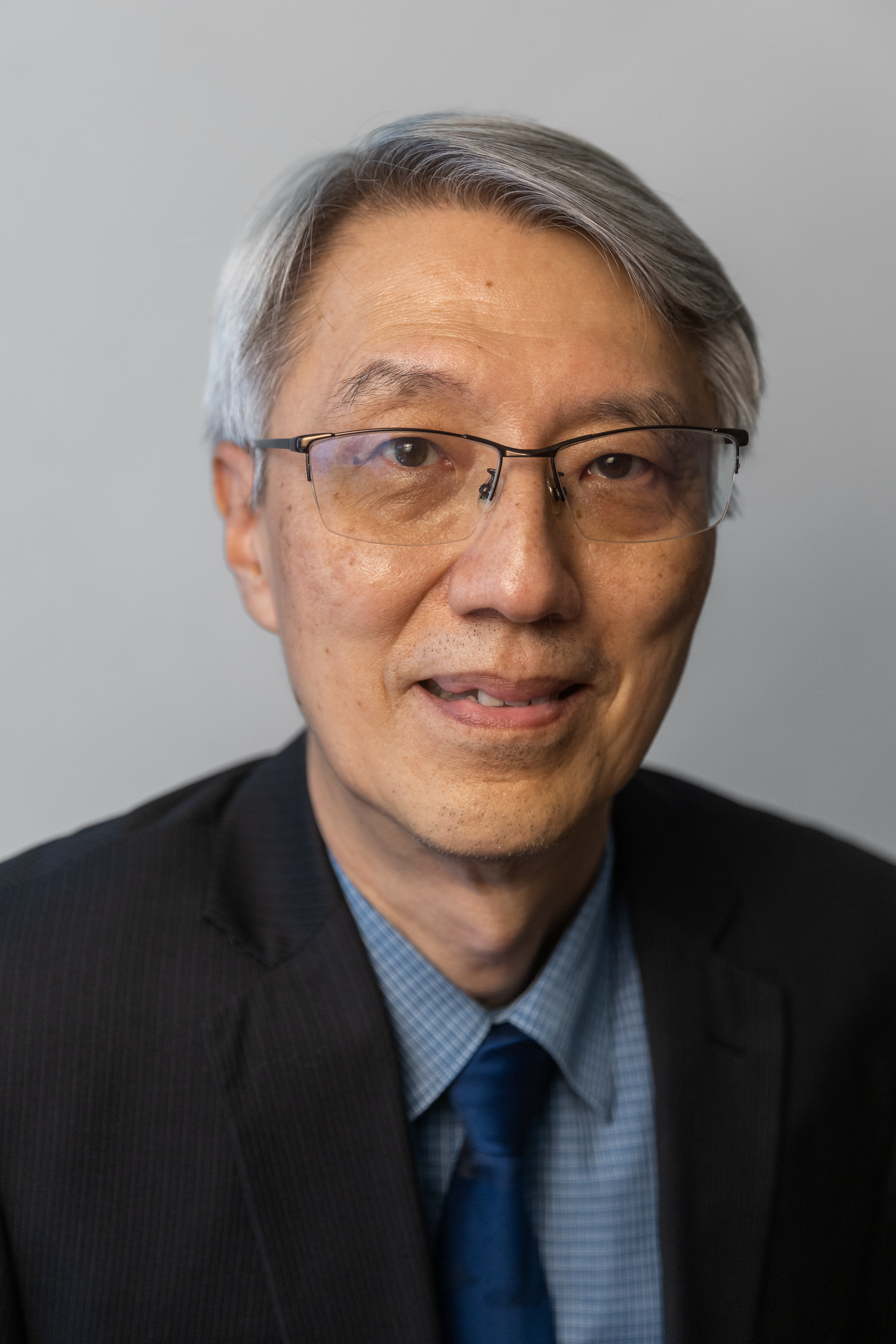}\\
\noindent
\textbf{Dennis C. H. Sng, PhD}\\
\textbf{Deputy Director \& Principal Scientist,  ROSE Lab}\\
\textbf{Nanyang Technological University (NTU),  Singapore}\\
Dr. Dennis Sng has been with the ROSE Lab at the Nanyang Technological University since 2013.  As Deputy Director, he is responsible for growing the Lab’s industry collaborations and its overall management.  As Principal Scientist, he is responsible for the research strategy and overall GPU computing systems infrastructure.  He also manages key industry funded research projects at the Lab. 
 
Dennis has almost 40 years of professional experience in research, government and industry.  He has a Ph.D. in Technology Management, a M.S. in Computer Science, and a B.Eng. in Electrical Engineering.\\
E-mail: dennis.sng@ntu.edu.sg

\hfill \break

\includegraphics[width=1in,height=1.25in,clip,keepaspectratio]{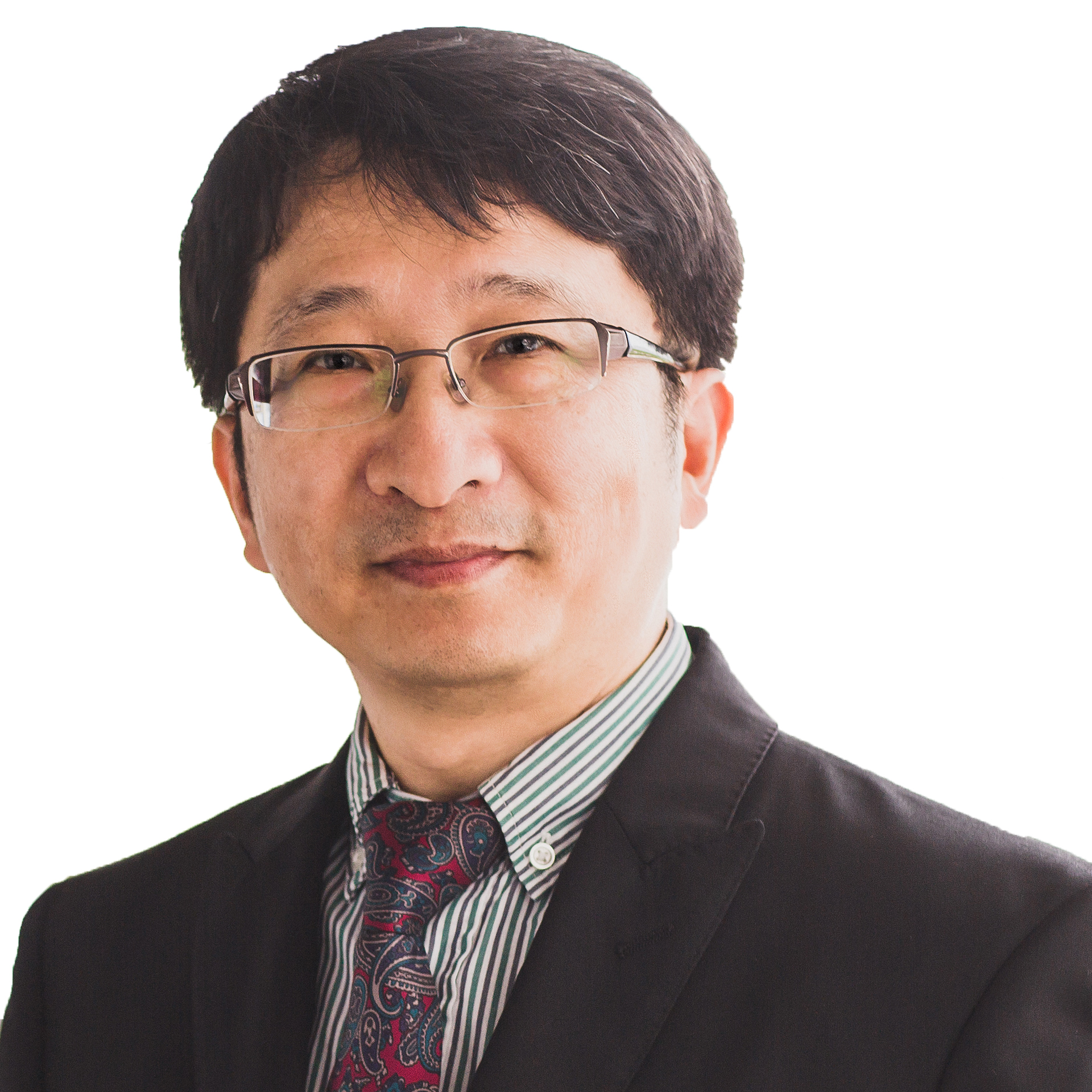}\\
\noindent
\textbf{Prof. Alex Kot} has been with the Nanyang Technological University, Singapore since 1991. He was Head of the Division of Information Engineering and Vice Dean Research at the School of Electrical and Electronic Engineering. Subsequently, he served as Associate Dean for College of Engineering for eight years. He is currently Professor and Director of Rapid-Rich Object Search (ROSE) Lab and NTU-PKU Joint Research Institute. He has published extensively in the areas of signal processing, biometrics, image forensics and security, and computer vision and machine learning.

Dr. Kot served as Associate Editor for more than ten journals, mostly for IEEE transactions. He served the IEEE SP Society in various capacities such as the General Co-Chair for the 2004 IEEE International Conference on Image Processing and the Vice-President for the IEEE Signal Processing Society. He received the Best Teacher of the Year Award and is a co-author for several Best Paper Awards including ICPR, IEEE WIFS and IWDW, CVPR Precognition Workshop and VCIP. He was elected as the IEEE Distinguished Lecturer for the Signal Processing Society and the Circuits and Systems Society. He is a Fellow of IEEE, and a Fellow of Academy of Engineering, Singapore.\\
E-mail: eackot@ntu.edu.sg

\newpage
\onecolumn
\begin{appendices}
\section{Additional Frequency Filtering Examples}\label{apx:freq}
\begin{figure}[htbp!]
\centering
\includegraphics[width=0.9\textwidth]{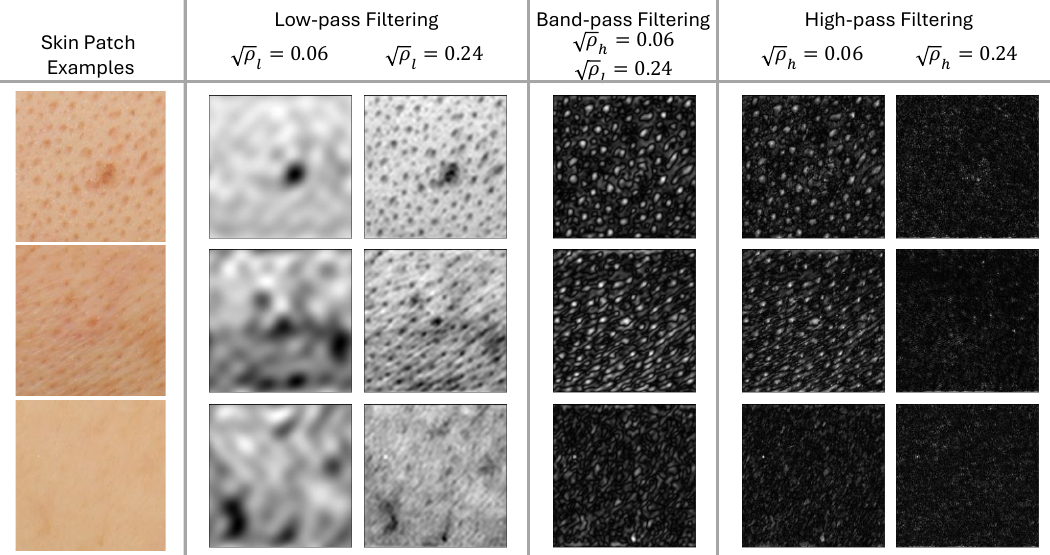}
\caption{Visual comparison of different frequency filtering. Removing too many high-frequency components results in a blurred image, while removing too many low-frequency components highlights only the finest skin features. In the case of low-pass filtering with $\sqrt{\rho_l}=0.24$, the images appear blurred, with some fine details smoothed out. On the other hand, high-pass filtering with $\sqrt{\rho_h}=0.06$ retains not only the texture details but also introduces high-frequency noise. Therefore, using a band-pass filter provides moderate skin texture details without overly emphasizing or diminishing them.}\label{fig:filter_examples}
\end{figure}

\section{Lighting Augmentation Examples}\label{apx:light}
\begin{figure}[htbp!]
\centering
\includegraphics[width=.9\textwidth]{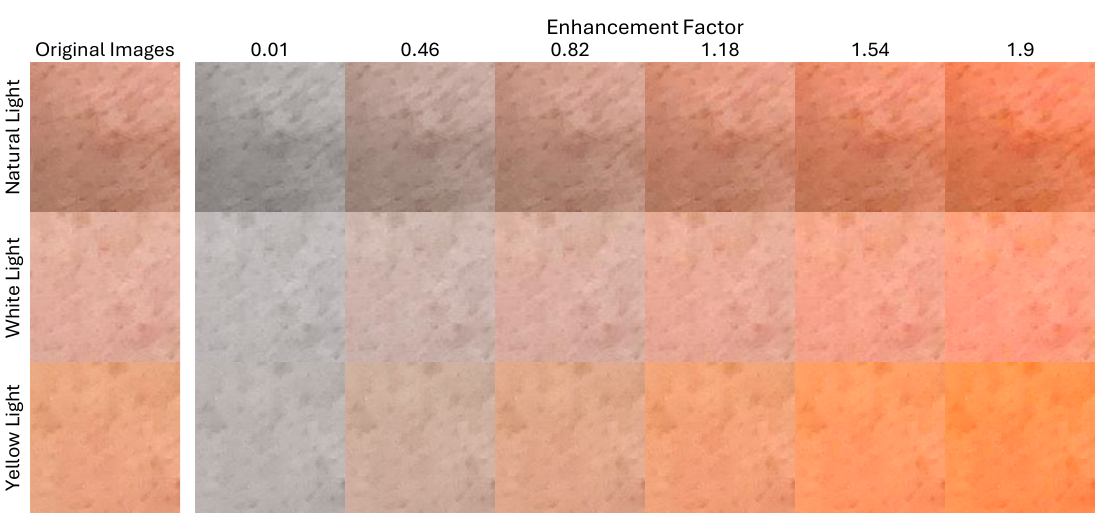}
\caption{Examples demonstrating the effects of varying enhancement factors using PIL.ImageEnhance.Color on skin patches captured under different lighting conditions. }\label{fig:pilcolor}
\end{figure}

\begin{figure}[htbp!]
\centering
\includegraphics[width=.9\textwidth]{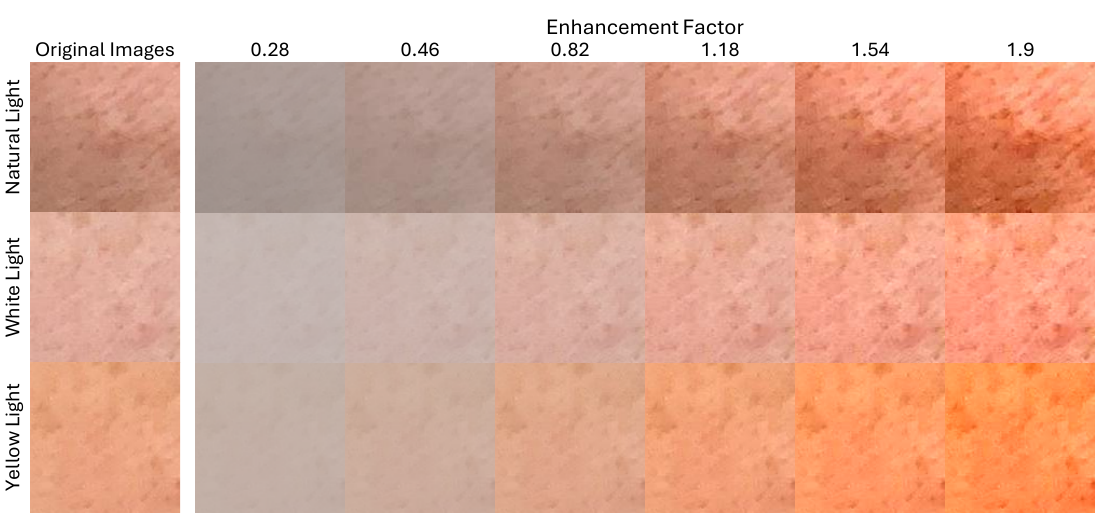}
\caption{Examples demonstrating the effects of varying enhancement factors using PIL.ImageEnhance.Contrast on skin patches captured under different lighting conditions.}\label{fig:pilcontrast}
\end{figure}

\begin{figure}[htbp!]
\centering
\includegraphics[width=.9\textwidth]{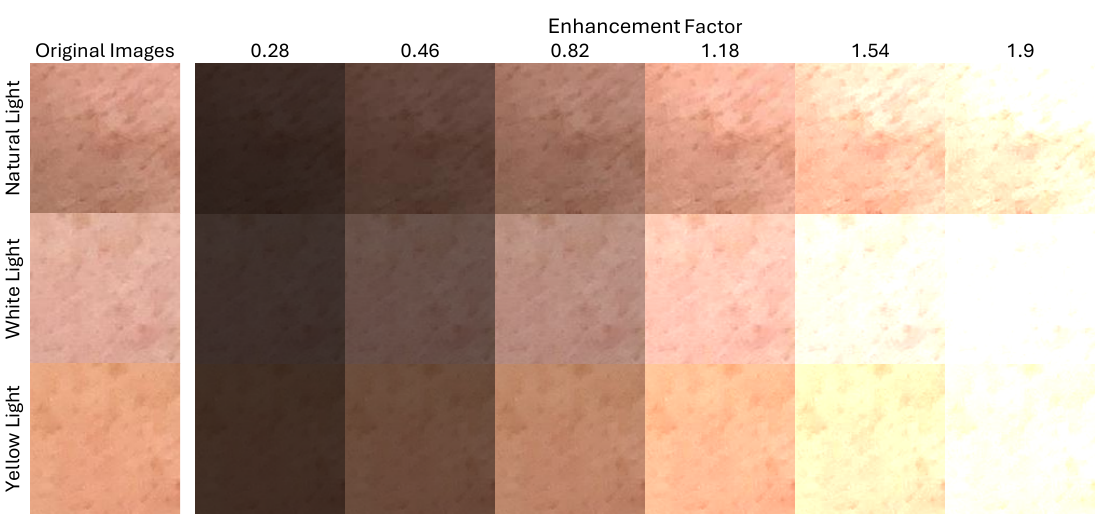}
\caption{Examples demonstrating the effects of varying enhancement factors using PIL.ImageEnhance.Brightness on skin patches captured under different lighting conditions.}\label{fig:pilbrightness}
\end{figure}

\begin{figure}[htbp!]
\centering
\includegraphics[width=.9\textwidth]{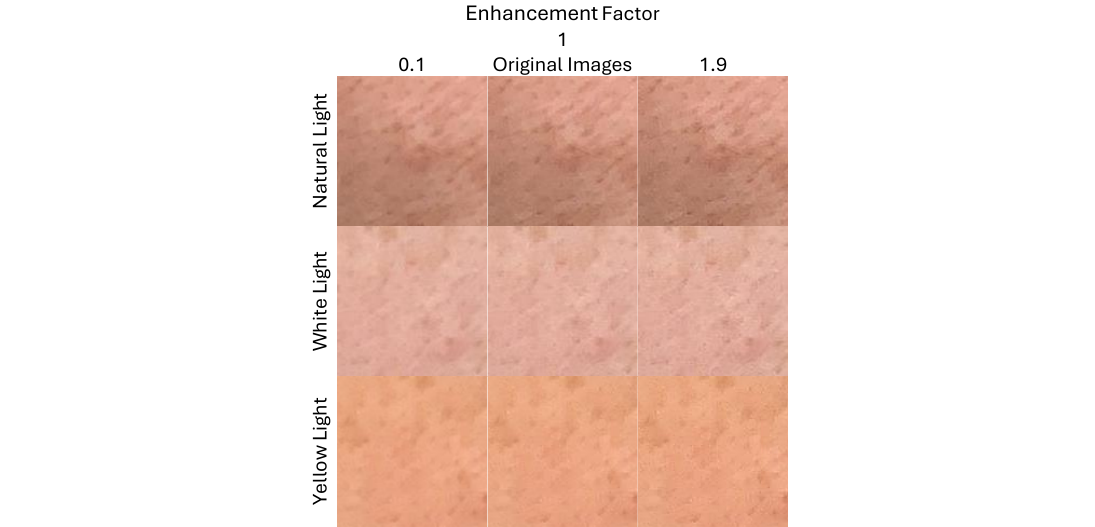}
\caption{Examples demonstrating the effects of varying enhancement factors using PIL.ImageEnhance.Sharpness on skin patches captured under different lighting conditions.}\label{fig:pilsharpness}
\end{figure}

\end{appendices}

\end{document}